\documentclass[lettersize,journal]{IEEEtran}
\usepackage{amsmath}
\usepackage{amssymb}
\usepackage{microtype}
\usepackage{graphicx}
\usepackage{svg}
\usepackage{subcaption}
\usepackage{caption}
\usepackage{tikz}
\usepackage{booktabs} 
\usepackage{multirow}
\usepackage{pgffor}
\usepackage{graphbox}
\usepackage{enumitem}

\usepackage{hyperref}
\usepackage{color}
\usepackage{xcolor,colortbl}
\usepackage{array}
\usepackage{pdfpages}

\usepackage[outdir=./]{epstopdf}
\usepackage{dsfont}

\usepackage[linesnumbered,ruled, vlined]{algorithm2e}
\usepackage{algorithmic}

\usepackage{natbib}
\usepackage{subcaption}

\usepackage{amsmath,amsfonts,bm}

\usepackage{amsthm}
\newtheorem{Problem}{Problem}

\def\eqref#1{equation~\ref{#1}}

\def\1{\bm{1}}

\DeclareMathAlphabet{\mathsfit}{\encodingdefault}{\sfdefault}{m}{sl}
\SetMathAlphabet{\mathsfit}{bold}{\encodingdefault}{\sfdefault}{bx}{n}

\newcommand{\std}[1]{\scriptsize{$\pm$#1}}
\newcommand{\modelname}{{TimeX++}} 

\newcommand{\oursa}{TimeX$_{a}$++}
\newcommand{\ourscf}{TimeX$_{cf}$++}

\definecolor{myblue}{HTML}{D7E5FA}
\definecolor{myblue1}{HTML}{D2EEED}
\definecolor{myblue2}{HTML}{AFE1E5}
\definecolor{mygreen}{HTML}{D2E7D1}
\definecolor{mygreen1}{HTML}{D4E6D4}
\definecolor{mygreen2}{HTML}{E0F6F1}
\definecolor{myyellow}{HTML}{FEF1CC}
\definecolor{myred}{HTML}{F6CCCA}
\definecolor{myred1}{HTML}{F5E8E3}
\definecolor{myred2}{HTML}{F7CCCA}
\definecolor{mypurple}{HTML}{E1D5E7}

\begin{document}
\title{Towards A Unified Information Bottleneck Framework for Time Series Explanations}

\author{%
Xu Zheng*, Zichuan Liu*, Zhuomin~Chen, Mayur Akewar, Janki Bhimani, Jason Liu, Mo Sha, Jingchao~Ni,\\ Wei Cheng, Dongsheng~Luo 
\IEEEcompsocitemizethanks{%
    \IEEEcompsocthanksitem X. Zheng, Z. Chen, M. Akewar, J. Bhimani, J. Liu, and M. Sha are with Florida International University.
    E-mail: \{xzhen019, zchen051, makew001, jbhimani, liux, msha \}@fiu.edu  
    \IEEEcompsocthanksitem  Z. Liu is with Carnegie Mellon University.  
    E-mail: zichuanl@andrew.cmu.edu
    \IEEEcompsocthanksitem J. Ni is with the University of Houston.
    E-mail: jni7@uh.edu%
    \IEEEcompsocthanksitem W. Cheng is with NEC Labs America.
    E-mail: weicheng@nec-labs.com%
    \IEEEcompsocthanksitem D. Luo is with Singapore Management University. This work was partially conducted while he was at Florida International University.
    E-mail: luodongsheng01@gmail.com%
    \IEEEcompsocthanksitem * indicates equal contribution.
}}%

\markboth{IEEE Transactions on Pattern Analysis and Machine Intelligence, VOL. X, NO. X, JULY 26}%
{\MakeLowercase{\textit{XXX et al.}}: Addressing Distribution Shift of Explanations for Time Series}

\IEEEtitleabstractindextext{%
\begin{abstract}
Explaining deep learning models operating on time series data is crucial in various applications that require transparent and interpretable insights into model behavior. {Existing explanation methods generally fall into two categories: attribution-based explanations, which identify the temporal regions most responsible for a prediction, and counterfactual explanations, which reveal how an input should be modified to alter the model's decision.}
{Despite valuable insights, these two fields are largely studied independently. This disconnect leaves attribution methods lacking causal validation, while counterfactual methods suffer from severe instability, producing adversarial-like noise instead of meaningful explanations.}
In this work, we revisit time-series explainability from an information-theoretic perspective and show that existing explainers are vulnerable to trivial solutions and distributional shifts. 
To address these limitations, we propose a unified objective function for explainable time series learning that bridges attribution and counterfactual reasoning within a single framework.
Building upon the Information Bottleneck principle, our formulation explicitly prevents trivial explanations and out-of-distribution counterfactuals.  
{Based on this objective function, we introduce {\modelname}, a novel explanation framework that learns a parametric transformation network to construct explanation-embedded instances, where preserved information yields attribution explanations and controlled information removal produces stable counterfactual explanations.}
We evaluate {\modelname} on synthetic and real-world benchmarks against state-of-the-art baselines. Extensive quantitative and qualitative results show that {\modelname} consistently outperforms competing methods, yielding faithful attributions and stable counterfactual explanations.
\end{abstract}
}

\maketitle

\IEEEdisplaynontitleabstractindextext
\IEEEpeerreviewmaketitle

\section{Introduction}
\label{sec:intro}

Deep learning has become a cornerstone technology in analyzing time series data, prevalent in scenarios such as finance~\cite{bento2021timeshap}, healthcare~\cite{kaushik2020ai}, and environmental science~\cite{zheng2026uncovering}. Despite its predictive success, the black-box nature of deep neural networks remains a critical limitation. The lack of explainability hinders user trust and prevents domain experts from gaining actionable insights, which are essential in these sensitive applications\cite{rudin2019stop,ghassemi2021false,ribeiro2016should}.

Current efforts to enhance the explainability of time series models primarily fall into two disconnected paradigms: attribution and counterfactual explanations. 
Attribution methods focus on a post-hoc pinpointing of the most salient temporal regions that dominate the model's prediction~\cite{ismail2020benchmarking,crabbe2021explaining,tonekaboni2020went}. Popular perturbation-based attribution methods, such as Dynamask~\cite{crabbe2021explaining} and Extrmask~\cite{enguehard23a}, evaluate feature importance by masking non-salient regions.  Nevertheless, these methods often rely on ad-hoc objectives lacking a solid theoretical foundation and frequently suffer from out-of-distribution (OOD) issues when the masked sub-instances are evaluated by the classifier~\cite{slack2020fooling, zhao2022ood, queen2023encoding}.
Conversely, counterfactual explanations answer ``what-if'' scenarios by finding the minimal perturbation required to change the model's prediction~\cite{chen2026signals}. Nevertheless, generating counterfactuals in the vast and continuous space without structural guidance often leads to trivial solutions, such as mode collapse in conditional generation~\cite{pmlr-v180-nemirovsky22a, dice, Jeanneret_2022_ACCV}. Pursuing label changes alone can result in adversarial attacks that satisfy the classifier's decision boundary while failing to capture meaningful and interpretable changes~\cite{laugel2019dangers, freiesleben2022magic,pmlr-v151-pawelczyk22a}. 

Crucially, existing research typically treats attribution and counterfactual explanations as completely isolated tasks, which introduces fundamental flaws~\cite{mothilal2021towards}. When treated separately, attributions remain purely descriptive, highlighting salient features but not causally verifying that altering those specific features actually changes the model's decision~\cite{janzing2020feature, miller2019explanation}.  Conversely, without the structural guidance provided by attributions, counterfactual generation lacks semantic grounding. If a counterfactual method optimizes for a label flip across the entire time series without knowing which features are semantically responsible for the original prediction, it inevitably produces adversarial noise rather than a valid, interpretable pattern shift.

\begin{figure*}[t]
\centering
\includegraphics[width=1.0\textwidth]{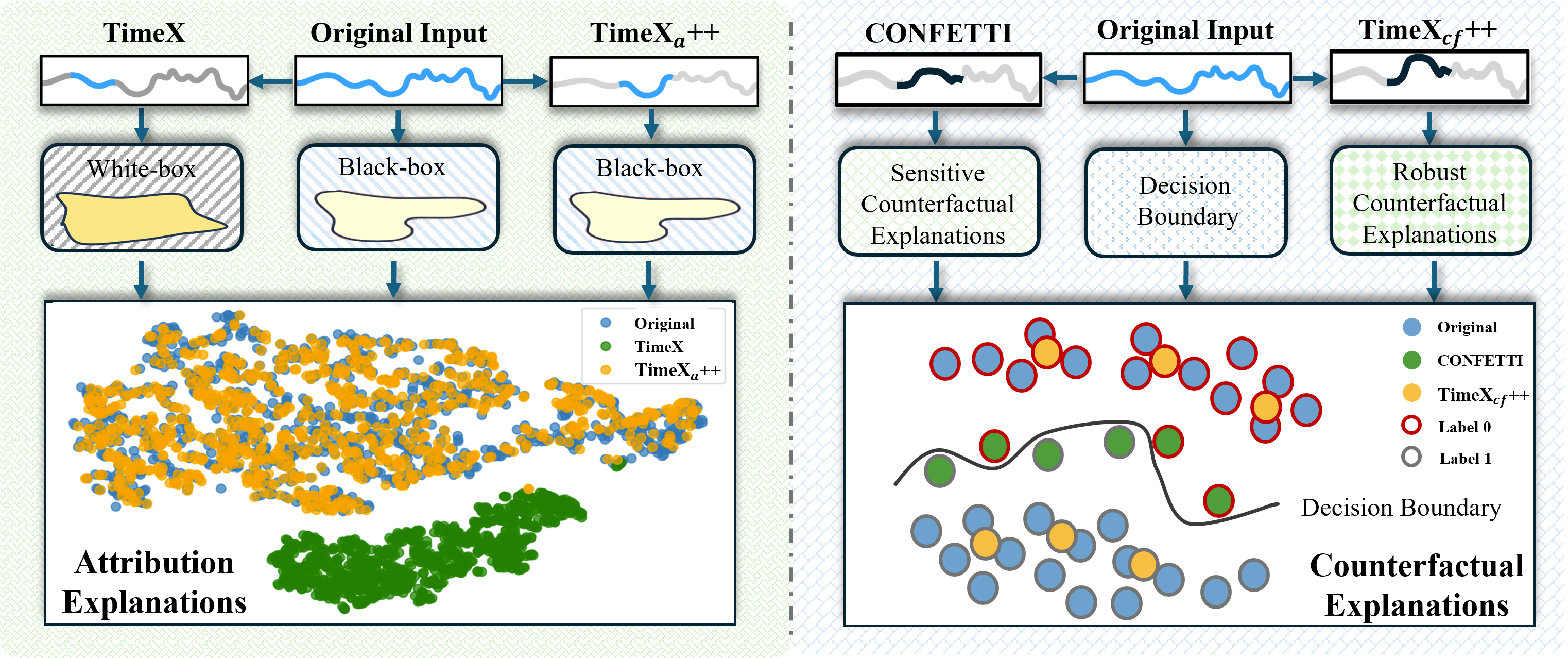}
\caption{A comparison between our model and previous work. For the attribution explanations, the latent embeddings of explanations are learned from the ECG dataset. Our explanations are within the original distribution, whereas the reference model's explanations are not. For the counterfactual explanations, our framework represents more robust results.}
\label{intro}
\vspace{-4mm}
\end{figure*}
To overcome these limitations, we establish a theoretical foundation that unifies attribution and counterfactual explanations through the lens of the Information Bottleneck (IB) principle~\cite{tishby2015deep}. Formally, given a time series instance $X$ and its label $Y$, the IB principle formulates the extraction of an \textit{explanation} sub-instance $X'$ as an optimization problem that balances compactness and informativeness: $X'$ minimizes $I(X; X')-\alpha I(X'; Y)$, where $I(\cdot;\cdot)$ denotes mutual information and $\alpha>0$ controls the trade-off~\cite{miao2022interpretable}.
In this context, the \textbf{attribution explanation} equates exactly to solving this IB objective. By minimizing $I(X; X')$, it discards redundant temporal dynamics, and by maximizing $I(X'; Y)$, it retains the sufficient statistics necessary for predicting $Y$. Consequently, the attribution $X'$ acts as the optimal information bottleneck, defining a semantically salient sub-manifold of the original high-dimensional temporal space.
 
Building upon this, we analyze the \textbf{counterfactual explanation} within the same information-theoretic framework. Generating a counterfactual requires finding a perturbed instance $X''$ that alters the classifier's prediction to a target label $Y''$. 
However, optimizing this perturbation directly in the unconstrained space of $X$ inevitably explores the whole decision space, which could cause injecting adversarial noise into the redundant regions $X \setminus X'$ that the IB explicitly discarded. 
To ensure a semantically valid distribution shift, the counterfactual perturbation must be strictly bounded within the support of the information bottleneck $X'$. From an information-theoretic perspective, a legitimate counterfactual transition $Y \rightarrow Y''$ necessitates a corresponding manipulation within the sufficient statistics subspace defined by $I(X'; Y)$, rather than arbitrary deviations in the redundant space. Through this IB-driven formulation, attribution and counterfactual explanations are intrinsically unified as complementary mechanisms over the same information bottleneck. The attribution identifies the exact semantic support $X'$, while the counterfactual manipulation within this bounded support provides the causal verification that $X'$ indeed dictates the predictive behavior.

However, directly applying the standard IB principle to time series presents critical challenges. Primarily, the exact computation of mutual information in continuous, high-dimensional temporal spaces is computationally intractable~\cite{mcallester2020formal}. Furthermore, attempting to evaluate the extracted sub-instances $X'$ directly with the original classifier often violates the underlying data manifold, inevitably leading to OOD predictions and unreliable evaluations~\cite{hooker2019benchmark,ICLR2025_2231d1ab}.
To address these challenges, we propose \modelname, a novel and unified framework designed to simultaneously extract both attribution and counterfactual explanations. It optimizes a practical, unified objective function. 
We replace the compactness quantifier $I(X; X')$ with a tractable variational upper bound comprising minimality and discrete constraints, and substitute the informativeness quantifier $I(X'; Y)$ with a measure of label consistency that strictly preserves the underlying data distribution. The key advantage of our unified architecture lies in its dual capability. First, for attribution explanations, it generates in-distribution, explanation-embedded instances, entirely circumventing the severe OOD issues that plague reference models (as illustrated in Figure~\ref{intro}). Second, for counterfactual explanations, by tightly coupling the counterfactual search with the in-distribution attribution bottleneck, \modelname~constrains the counterfactual trajectory along meaningful feature dimensions. This guarantees the generation of stable, semantically valid counterfactuals that are highly resistant to noise vulnerabilities.
We summarize our contributions as follows:
\begin{itemize}[leftmargin=*]
\item We formally investigate the limitations of treating attribution and counterfactual explanations as isolated tasks in time series. By analyzing them through the Information Bottleneck principle, we establish their theoretical connection and propose a practical, unified objective function.
\item We propose a novel explanation framework, \modelname, which successfully addresses the out-of-distribution shifting issue in attribution explanations by generating in-distribution sub-instances and leveraging this attribution bottleneck to generate stable, semantically grounded counterfactuals.
\item We achieve state-of-the-art performance for both attribution and counterfactual explanation tasks on synthetic and real-world datasets and demonstrate their effectiveness.
\end{itemize}

\section{Related Work}
\label{sec:related}

\paragraph{Attribution Explanations in Time Series}
Existing explainable AI (XAI) methods for time series primarily focus on feature attribution, which aims to identify the most salient temporal regions dictating a well-trained model's decision~\cite{enguehard23a, bento2021timeshap}. While early approaches relied on attention weights~\cite{choi2016retain} or gradient signals~\cite{sundararajan2017axiomatic}, perturbation-based methods have become the dominant paradigm. 
These methods evaluate feature importance by masking or altering data segments using static baselines~\cite{suresh2017clinical}, generative models~\cite{tonekaboni2020went}, or information diminution~\cite{crabbe2021explaining, liu2024explaining}. However, a critical flaw in perturbation-based attribution is the OOD problem that when salient features are masked or substituted, the resulting sub-instances often violate the underlying data manifold~\cite{hooker2019benchmark}. Consequently, querying the original classifier with these OOD samples leads to unreliable evaluations~\cite{zhao2022ood, slack2020fooling}.

\paragraph{Counterfactual Explanations for Time Series}
In contrast to descriptive attribution, counterfactual explanations seek minimal modifications to an input required to alter the model's prediction. For time series data, existing paradigms are broadly classified into three categories. First, replacement-based methods (e.g., CoMTE~\cite{COMTE}, Native Guide~\cite{delaney2021instance}) substitute contiguous segments using nearest unlike neighbors or temporal shapelets~\cite{bahri2022shapelet, MGCF}. Second, gradient-based approaches (e.g., LatentCF++~\cite{wang2021learning}, Glacier~\cite{wang2024glacier}) directly optimize perturbations in the input or learned latent spaces. Third, genetic frameworks (e.g., TSEvo~\cite{hollig2022tsevo}, Sub-SpaCE~\cite{refoyo2024sub}) explore the vast space of temporal segment edits using evolutionary objectives. Despite these advances, generating robust time series counterfactuals remains challenging. Because continuous temporal space is unconstrained, counterfactual optimization often exploits local decision boundary vulnerabilities. This results in the injection of meaningless noise that acts as an adversarial attack rather than providing a semantically interpretable pattern shift~\cite{laugel2019dangers, freiesleben2022magic}.

\section{Problem Formulation} 
\label{sec:notations}

This work focuses on explainability in time series classification. Let $X \in \mathbb{R}^{T\times D}$ 
be a multivariate time series instance of length $T$ with $D$ features, and $\mathcal{X}$ is input space. A multivariate time series is one for which $D>1$, or univariate. The value of the feature indexed $d$ at time $t$ is denoted by $X[t,d]$.  A training set  $\mathcal{T}=\{(X_i,Y_i)|i\in [N]\} $ consists of $N$ time series instances $X_i$ along with their associated labels $Y_i$, where $Y_i\in \mathcal{C}$ and $\mathcal{C}=\{1,2,\cdots,|\mathcal{C}|\}$ is the label space.
A pre-trained black-box classifier $f: \mathcal{X} \rightarrow \mathcal{C}$ maps $X$ to a predicted label $Y \in \mathcal{C}$. 

Our objective is to formally unify feature attribution and counterfactual generation under the IB principle. 
In this unified view, the attribution acts as the optimal information bottleneck, and the counterfactual acts as a semantic perturbation strictly bounded within that bottleneck.  
To develop a general explainability framework, we focus on \textit{post-hoc, instance-level} methods that are task-agnostic and treat the model $f(\cdot)$ as a black box~\cite{zhang2021survey}. Furthermore, we study model explainability, where the extracted sub-instance is required to be sufficient for the model output rather than the ground-truth label~\cite{faber2021comparing, liu2024explaining}.

\begin{Problem}[Attribution Explanations]
\label{prob:attr}
Given a trained model $f$ and input $X$, the objective in post-hoc instance-level time series explanation is to find a sub-instance $X'$ that `explains' the prediction of $f$ on $X$.   The sub-instance $X'$ is obtained by applying a binary mask $M \in \mathcal{M} =\{0,1\}^{T\times D}$ on $X$, i.e.,  $X' = X \odot M$, where $\odot$ is the element-wise multiplication.
\end{Problem}

To transition this discrete extraction into a differentiable continuous optimization, we define an attribution extractor $g(\cdot)$ that maps the input $X$ to a mask $M \in [0,1]^{T\times D}$. Each element is independently sampled from a Bernoulli distribution $\pi_{t,d}$.

\begin{Problem}[Counterfactual Explanations]
\label{prob:cf}
Given a trained model $f$, an input time series $X$ with predicted label $Y=f(X)$, and a target label $Y'' \in \mathcal{C}$ ($Y'' \neq Y$), the objective is to find a counterfactual instance $X''$ such that $f(X'') = Y''$, while minimizing the perturbation distance $d(X, X'')$, where the counterfactual search space is restricted to the semantic support set defined by the attribution mask $M$.
\end{Problem}

\section{Explaining Time Series Learning via Information Bottleneck}
\label{sec:objective}
The IB principle extracts an attribution explanation sub-instance $X' = X \odot M$ by maximizing $I(X'; Y) - \alpha I(X; X')$. However, directly estimating the mutual information $I(X'; Y)$ for informativeness in high-dimensional continuous spaces is computationally intractable. To overcome and extend this, following established practices in explainable learning (e.g., PGExplainer~\cite{luo2020parameterized}), we substitute the informativeness term with a tractable Label Consistency ($\mathrm{LC}$) measure (i.e., cross-entropy), yielding the modified IB formulation:
\begin{equation}
    \label{eq:modified_ib}
    \begin{aligned}
        \min_{M \sim \mathrm{Bern}(\bm{\pi})} -\mathrm{LC}\big(Y^{expl}, f(\tilde{X})\big) + \alpha I(X; X'),
    \end{aligned}    
\end{equation}
where $Y^{expl}$ is the objective explanation label, and $\tilde{X}$ is the explanation ($X'$ for attribution and $X''$ for counterfactual explanation). 
While Eq.~(\ref{eq:modified_ib}) bypasses the informativeness estimation, directly applying it to time series explainability introduces three critical interconnected problems. \textbf{First}, regarding the intractability of compactness, estimating the mutual information $I(X; X')$ remains computationally prohibitive and suffers from severe sample complexity bounds in high-dimensional temporal spaces. \textbf{Second}, a severe OOD problem arises during attribution evaluation. When $Y^{expl}$ equals $Y$, computing $f(X')$ requires passing a discretely masked instance directly to the classifier. This drastically violates the underlying data manifold, causing OOD predictions and destroying the reliability of the $\mathrm{LC}$ term~\cite{hooker2019benchmark}. \textbf{Third}, counterfactual generation suffers from adversarial instability. When $Y^{expl}$ is set to a target label $Y''$ to seek a perturbed instance $X''$, unconstrained optimization of $\mathrm{LC}$ over the continuous temporal space inevitably exploits the classifier's vulnerabilities. This generates meaningless adversarial noise rather than semantically valid shifts~\cite{freiesleben2022magic}. To resolve these interconnected problems, we reconstruct the objective into a unified framework that enforces distribution consistency and structural stability.

\begin{figure*}[t]
\centering 
\includegraphics[width=1.0\textwidth]{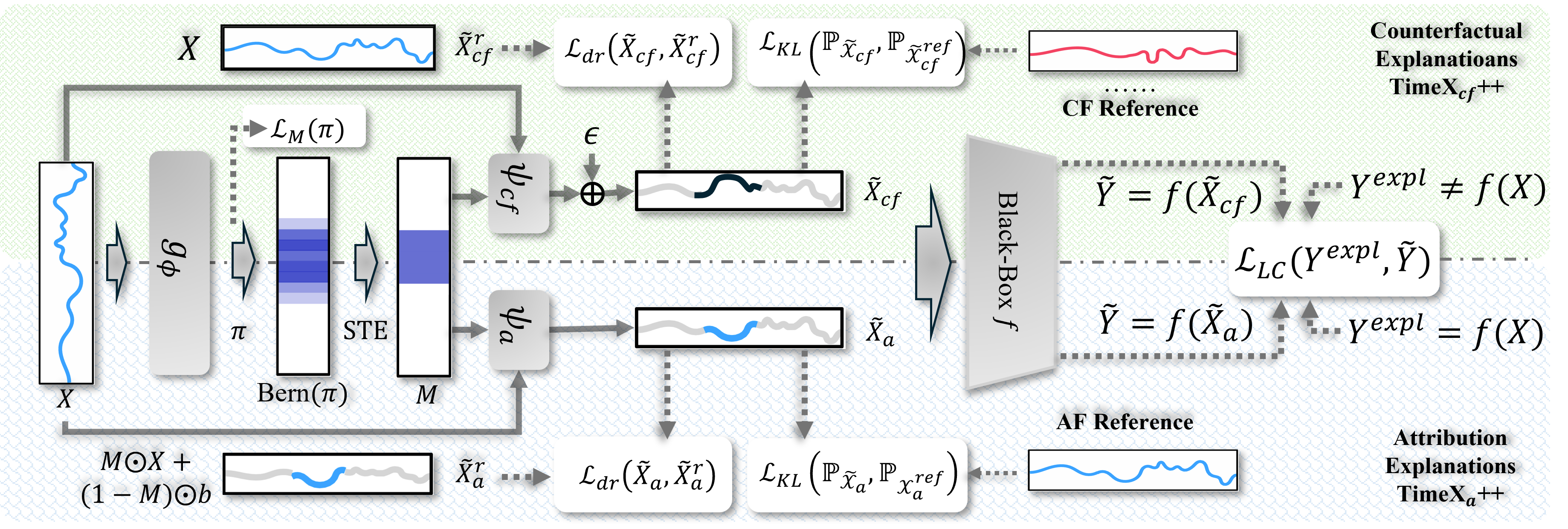}
\caption{ {The overall architecture of \modelname, a unified framework for attribution and counterfactual explanations supported by Information Bottleneck. ~{\oursa} and ~{\ourscf} have similar components and loss functions but different implementations.} 
}\label{framwork}
\vspace{-2mm}
\end{figure*}

\subsection{Tractable Compactness Bound}
To bypass the intractable mutual information $I(X;X')$, we replace it with a tractable variational upper bound. Since the physical goal of minimizing $I(X;X')$ is to enforce sparsity and determinism in the explanation bottleneck, we directly penalize the expected size $|M|$ and the entropy $H(M)$ of the mask. By defining a prior Bernoulli distribution $\mathbb{Q}(M)$ with a parameter $r$, we analytically bound the compactness term as a Kullback-Leibler (KL) divergence between the posterior mask distribution $\mathbb{P}(M|X)$ and the prior:
\begin{equation}
    \label{eq:kl_bound}
    \begin{aligned}
    \mathcal{L}_{compact} = \mathbb{E}_{X} \left[ D_\mathrm{KL}(\mathbb{P}(M|X) \| \mathbb{Q}(M)) \right]
    \end{aligned}
\end{equation}
Minimizing this term efficiently compresses the bottleneck without requiring complex mutual information estimators.

\subsection{Distribution-Preserving and Bounded Generation}
To simultaneously solve the attribution OOD problem and the counterfactual instability problem, we project the bottleneck into an \textit{in-distribution} generated instance $\tilde{X} \in \mathcal{X}$, and evaluate $f(\tilde{X})$. The modified informativeness objective becomes:
\begin{equation}
    \label{eq:info_stability}
    \begin{aligned}
    \mathcal{L}_{info} = & \mathcal{L}_{\mathrm{LC}}\big(Y^{expl}, f(\tilde{X})\big) + \beta \mathcal{L}_{\mathrm{KL}}(\mathbb{P}_{\tilde{X}} \|  \mathbb{P}_{\mathcal{X}}) \\
    & + \lambda \mathcal{L}_{bound}(\tilde{X}, X, M),
    \end{aligned}
\end{equation}
where $\mathcal{L}_{\mathrm{LC}}(\cdot,\cdot)$ equals to $\mathrm{LC}(\cdot,\cdot)$. This introduces two crucial regularization terms. The distribution preservation penalty, $\mathcal{L}_{\mathrm{KL}}(\mathbb{P}_{\tilde{X}} \| \mathbb{P}_{\mathcal{X}})$, explicitly regularizes the distribution shift between the generated $\tilde{X}$ and the true data manifold $\mathcal{X}$. This mathematically eliminates the OOD problem, ensuring the classifier evaluates valid, realistic samples. Concurrently, the bottleneck constraint, $\mathcal{L}_{bound}(\tilde{X}, X, M)$, acts as the causal structural anchor. It forces the generated perturbation $(\tilde{X} - X)$ to be strictly bounded by the spatial and temporal active regions of the attribution mask $M$. By penalizing changes in the redundant background, this constraint fundamentally prevents unconstrained adversarial search.

\subsection{Unified Architecture Integration}
Combining the tractable compactness bound and the regularized generation, the final unified objective of \modelname~is:
\begin{align*}
    \min_{M, \tilde{X}} \; \mathcal{L}_{info} + \alpha \mathcal{L}_{compact} .
\end{align*}
This single cohesive architecture systematically integrates attribution and counterfactual explanations through the shared bottleneck $M$. For \textbf{attribution} extraction, we set $Y^{expl} = Y$. The optimization minimizes $\mathcal{L}_{compact}$ to find the tightest possible mask $M$, while $\mathcal{L}_{\mathrm{KL}}$ ensures the generated evaluation instance $\tilde{X}$ remains strictly in-distribution, resolving the OOD problem. Conversely, for \textbf{counterfactual} generation, we set $Y^{expl} = Y''$. Although the optimization seeks to flip the label, $\mathcal{L}_{bound}$ explicitly locks the counterfactual trajectory within the semantic support of $M$. 
This guarantees that the counterfactual only manipulates the causal features identified by the attribution, resolving the adversarial instability problem. Through this integration, \modelname~provides causally verified attributions and semantically grounded counterfactuals.

\section{The \modelname~Unified Framework}
\label{sec:framework}
Translating the tractable unified Information Bottleneck (IB) objective from Section~\ref{sec:objective} into a practical neural architecture, we introduce the \modelname~framework (depicted in Figure~\ref{framwork}). The framework is constructed upon two primary neural modules working in tandem: the shared \textbf{Attribution Bottleneck Extractor ($g_\phi$)} and the paradigm-specific \textbf{Conditioned Generators} ($\psi_a$ and $\psi_{cf}$). By dynamically configuring the objective explanation label $Y^{expl}$ and the specific generator, \modelname~seamlessly instantiates into two operational modes: {\oursa}  and {\ourscf}.

\subsection{Attribution Bottleneck Extractor}
The extractor $g_\phi: \mathbb{R}^{T\times D } \mapsto [0,1]^{T\times D}$ serves as the shared semantic foundation for both explanation paradigms. It is implemented via an encoder-decoder Transformer that maps the input time series $X$ into a stochastic probability matrix $\bm{\pi}$.
To bridge the gap between continuous network optimization and the discrete mask $M$ required by the IB formulation, we employ the Straight-Through Estimator (STE)~\cite{jang2016categorical}. During the forward pass, we sample a deterministic binary mask $M = \text{STE}(\operatorname{Bern}(\bm{\pi})) \in \{0,1\}^{T\times D}$. During the backward pass, gradients bypass the discrete sampling operator, allowing the parameters $\phi$ to be optimized smoothly.

To satisfy the tractable compactness bound $\mathcal{L}_{compact}$ (Eq.~\ref{eq:kl_bound}) without requiring complex mutual information estimators, we apply two structural regularizations to the predicted probability $\bm{\pi}$. First, to enforce sparsity and determinism, we explicitly minimize the KL divergence against a sparse prior $\mathbb{Q}(M)$ parameterized by $r$. For attribution explanations ({\oursa}), we defaultly set $r=0.5$. For counterfactual generation ({\ourscf}), we enforce a strictly narrower bottleneck by default setting $r=0.1$, isolating the most critical causal features to restrict the counterfactual search space. Second, to ensure temporal continuity, we penalize irregular mask fragmentation across the time axis. Because true semantic features in time series typically occur as contiguous temporal segments rather than isolated anomalous spikes,  the continuity loss is formulated as:
\begin{equation}
\label{eq.con}
\begin{aligned}  
\mathcal{L}_{\text{con}} =  \frac{1}{T\times D}  \sum_{d=1}^D \sum_{t=1}^{T-1} \sqrt{\left(\pi_{t,d}-\pi_{t+1,d}\right)^2}.
\end{aligned}
\end{equation}
The total bottleneck extraction loss is defined as $\mathcal{L}_{M} = \mathcal{L}_{compact} + \lambda_{\text{con}} \mathcal{L}_{\text{con}}$, formulated as:
\begin{equation}
\begin{aligned}
\mathcal{L}_M = & \sum_{t,d} \left[ \pi_{t,d}\log\left(\frac{\pi_{t,d}}{r}\right) + (1-\pi_{t,d})\log\left(\frac{1-\pi_{t,d}}{1-r}\right) \right] \\
& + \lambda_{\mathrm{con}} \mathcal{L}_{\text{con}} .
\end{aligned}
\end{equation}
where the first summation~\cite{miao2022interpretable} expands the KL divergence $\mathcal{L}_{compact}$, and $\lambda_{\mathrm{con}}$ controls the temporal continuity penalty. 

\subsection{Conditioned Generators}
The conditioned generators translate the structural bottleneck $M$ into the generated explanation-embedded instance $\tilde{X} \in \mathcal{X}$. They are explicitly designed to resolve the OOD problem for attribution explanations (using $\psi_a$) and the robustness problem for counterfactual explanations (using $\psi_{cf}$).

\paragraph{{\oursa}} 
When evaluating feature attribution, directly passing the masked instance $X \odot M$ to the classifier destroys the temporal manifold, causing arbitrary OOD predictions. {\oursa} solves this by projecting the masked data back into the true distribution.
We first define a baseline reference $\widetilde{X}_a^r = X \odot M + b \odot (1 - M)$, where background noise $b$ is sampled from the dataset distribution $\mathbb{B}_\mathcal{X}$. The attribution generator $\psi_a$ then maps $[X, M]$ into a smoothed, generated instance $\tilde{X}_a = \psi_a(X, M)$.
To guarantee it resolves the OOD problem, we enforce distribution preservation via $\mathcal{L}_{\mathrm{KL}}(\mathbb{P}_{\tilde{X}_a} \| \mathbb{P}_{\mathcal{X}_a^r})$. To ensure the generator strictly focuses on smoothing the OOD artifacts rather than ungrounded features, we apply the bottleneck constraint $\mathcal{L}_{bound}^{a}$ as a Euclidean penalty against the reference:
\begin{equation}
\mathcal{L}_{bound}^{a} = \frac{1}{TD} \sum_{d=1}^D \sum_{t=1}^{T} \left\| \tilde{X}_a[t,d] - \widetilde{X}_a^r[t,d] \right\|^2
\end{equation}
The attribution is then validated by matching the original prediction: $\mathcal{L}_{\mathrm{LC}}\big(f(X), f(\tilde{X}_a)\big)$.

\begin{algorithm}[t]
    \caption{Training Pipeline of \modelname}
    \label{alg:unified}
	\begin{algorithmic}[1]
	\STATE 	\textbf{Input:} Training set $\mathcal{T}$, Trained classifier $f$, mode $\in \{	\text{A}, 	\text{CF}\}$, Target $Y''$ (if CF), Hyperparams $\{\alpha, \beta, r, \lambda_{\mathrm{con}}\}$
    \STATE 	\textbf{Init:} Extractor $g_{\phi}$, Generators $\psi_a$, $\psi_{cf}$, and Variation function $\psi_n$
	\FOR{$e = 1$ to $E$}
		\FOR{$i = 1$ to $N$}
		    \STATE $\bm{\pi}_i = g_{\phi}(X_i)$; Sample $M_i = \operatorname{STE}(\operatorname{Bern}(\bm{\pi}_i))$
            \IF{mode == $\text{A}$}
		        \STATE Set reference $\widetilde{X}_{a,i}^r = M_i\odot X_i + (1- M_i)\odot b$
		        \STATE Generate instance $\tilde{X}_{a,i} = \psi_a(X_i, M_i)$
                \STATE $Y^{expl} = f(X_i)$; $\tilde{X}_i = \tilde{X}_{a,i}$; $\widetilde{X}_i^r = \widetilde{X}_{a,i}^r$
            \ELSE
                \STATE Set reference $\widetilde{X}_{cf,i}^r = X_i$
                \STATE Predict perturbation $E_i = \psi_{cf}(X_i, M_i)$
                \STATE Sample $X^{ref}$ with label $Y''$, compute $\epsilon = \psi_n(X_i, M_i, X^{ref})$
                \STATE Generate instance $	\tilde{X}_{cf,i} = X_i + M_i \odot E_i + \epsilon$
                \STATE $Y^{expl} = Y''$; $\tilde{X}_i = \tilde{X}_{cf,i}$; $\widetilde{X}_i^r = \widetilde{X}_{cf,i}^r$
            \ENDIF
		\ENDFOR
        \STATE Compute compactness $\mathcal{L}_{M} = \mathcal{L}_{compact} + \lambda_{\text{con}} \mathcal{L}_{	\text{con}}$
        \STATE Compute bottleneck constraint $\mathcal{L}_{bound}$ against $\widetilde{X}^r$
        \STATE Compute distribution preservation $\mathcal{L}_{\mathrm{KL}}$
        \STATE Compute Label Consistency $\mathcal{L}_{\mathrm{LC}}(Y^{expl}, f(\tilde{X}))$
  		\STATE Total loss: $\widetilde{\mathcal{L}} = \mathcal{L}_{\mathrm{LC}} + \alpha \mathcal{L}_{M} + \beta(\mathcal{L}_{\mathrm{KL}} + \mathcal{L}_{bound})$
  		\STATE Update $\phi$ and active generators ($\psi_a$, or $\psi_{cf}, \psi_n$) via gradient descent
	\ENDFOR
	\end{algorithmic}  
\end{algorithm}

\paragraph{{\ourscf}}
Unconstrained counterfactual generation in continuous temporal space inevitably exploits classifier null spaces, producing adversarial noise. {\ourscf} solves this robustness problem by strictly locking the perturbation within the semantic bottleneck $M$.
Instead of generating an entirely new instance, the counterfactual generator $\psi_{cf}$ learns a target-driven perturbation matrix $E = \psi_{cf}(X, M)$. The counterfactual instance is structurally formulated as:
\begin{equation}
\label{eq:cf_gen}
\tilde{X}_{cf} = X + M \odot E + \epsilon
\end{equation}
where $\epsilon$ introduces target-class reference noise. Specifically, we define an auxiliary variation function $\epsilon = \psi_n(X, M, X^{ref})$, which incorporates fine-grained target-class dynamics from a randomly sampled training instance $X^{ref}$ belonging to the target label $Y''$. 
During inference, this simplifies deterministically to $\tilde{X}_{cf} = X + M \odot E$.
Crucially, to mathematically alleviate adversarial vulnerabilities, the bottleneck constraint $\mathcal{L}_{bound}^{cf}$ heavily penalizes any deviation from the original reference ($\widetilde{X}_{cf}^r = X$) in the non-bottleneck background regions ($1-M$):
\begin{equation}
\begin{aligned}
& \mathcal{L}_{bound}^{cf}  = \\
& \frac{1}{TD} \sum_{d=1}^D \sum_{t=1}^{T} \left\|  (1-M[t,d]) \odot (\tilde{X}_{cf}[t,d] - \widetilde{X}_{cf}^r[t,d]) \right\|^2
\end{aligned}
\end{equation}
This formulation acts as a hard causal anchor, which prevents the model from injecting imperceptible adversarial noise into redundant features, forcing the targeted prediction shift $\mathcal{L}_{\mathrm{LC}}\big(\tilde{Y}, f(\tilde{X}_{cf})\big)$ to rely on robust semantic manipulations.

\subsection{End-to-End Optimization} 
The entire \modelname~framework is optimized end-to-end. By sharing the underlying architecture and unifying the mathematical formulation, both {\oursa} and {\ourscf} minimize the same overarching objective derived in Section~\ref{sec:objective}:
\begin{equation}
\label{eq:overall_loss}
\widetilde{\mathcal{L}} = \mathcal{L}_{\mathrm{LC}}\big(Y^{expl}, f(\tilde{X})\big) + \alpha \mathcal{L}_{M} + \beta \big( \mathcal{L}_{\mathrm{KL}} + \mathcal{L}_{bound} \big)
\end{equation}
This allows our method to seamlessly alternate between extracting robust attributions and generating causally constrained counterfactuals simply by toggling the objective label $Y^{expl}$ and the specific generator ($\psi_a$ or $\psi_{cf}$). The unified train procedure is summarized in Algorithm~\ref{alg:unified}.

\begin{table}[t]
\centering
    \setlength{\tabcolsep}{3pt}
    \caption{The description of synthetic and real-world datasets. }
    \label{tab:dataset}
    \resizebox{1\columnwidth}{!}{
    \begin{sc}
        \begin{tabular}{c|cccc}
        \toprule
        \textbf{Dataset} &  \textbf{\# of Samples} & \textbf{Length} & \textbf{Dimension} & \textbf{Classes} \\ 
        \midrule
        FreqShapes & 6,100 & 50 & 1 & 4 \\
        \rowcolor{gray!15} SeqComb-UV & 6,100 & 200 & 1 & 4 \\
        SeqComb-MV & 6,100 & 200 & 4 & 4 \\
        \rowcolor{gray!15} LowVar & 6,100 & 200 & 2 & 4 \\
        \midrule
        ECG & 92,511 & 360 & 1 & 2  \\
        \rowcolor{gray!15} PAM &  5,333 & 600 & 17 & 8  \\
        Epilepsy & 11,500 & 178 & 1 & 2   \\
        \rowcolor{gray!15} Boiler & 160,719 & 36 & 20 & 2   \\
        Wafer & 7,164 & 152 & 1 & 2 \\
        \rowcolor{gray!15} FreezerRegular & 3,000 & 301 & 1 & 2 \\
        \bottomrule
        \end{tabular}
    \end{sc}
    }
    \vspace{-2mm}
\end{table}

\section{Experiments}
\label{sec:experiments}

In this section, we empirically evaluate the proposed \modelname~framework. By unifying attribution and counterfactual explanations through the lens of the Information Bottleneck, we aim to demonstrate three key aspects of their effectiveness. 
\textbf{First}, we assess {\modelname}'s ability to generate faithful attribution explanations and effective counterfactual explanations.
\textbf{Second}, we examine the extent to which the proposed distribution-matching constraint alleviates the out-of-distribution shift. 
\textbf{Third}, we evaluate the robustness of counterfactual explanations and investigate whether \modelname~ effectively guides counterfactual generation toward semantically meaningful outcomes, preventing it from degenerating into trivial adversarial perturbations.

\subsection{Experimental Setup}
\label{sec:exp_setup}
To rigorously evaluate our framework, we establish a comprehensive experimental setup encompassing diverse datasets, established baselines, and strict evaluation metrics.

\subsubsection{Datasets}
We utilize ten datasets encompassing both synthetic and real-world time-series scenarios, adopting the standard configurations from established benchmarks~\cite{queen2023encoding}. The synthetic datasets, including FreqShapes, SeqComb-UV, SeqComb-MV, and LowVar, are meticulously crafted with inherent ground-truth explanations to encapsulate temporal dynamics, guarding against heuristic learning shortcuts~\cite{geirhos2020shortcut}. The real-world datasets cover various domains, such as electrocardiogram classification (ECG)~\cite{Moody2001TheIO}, action recognition (PAM)~\cite{reiss2012introducing}, electroencephalogram analysis (Epilepsy)~\cite{andrzejak2001indications}, mechanical fault detection (Boiler)\footnote{\url{https://dx.doi.org/10.21227/awav-bn36}}, and sensor classifications (Wafer, FreezerRegular)~\cite{dau2019ucr}. The Transformer~\cite{vaswani2017attention} is employed as the underlying black-box classifier across all tasks.

\subsubsection{Baselines}
For comprehensive comparative analysis, we benchmark \modelname~against eleven state-of-the-art explainers meticulously selected across both the attribution and counterfactual paradigms. For \textbf{attribution explanations}, we evaluate against six popular baselines: Integrated Gradients (IG)~\cite{sundararajan2017axiomatic} as a fundamental gradient-based explainer; Dynamask~\cite{crabbe2021explaining} and WinIT~\cite{leung2023temporal} as recent occlusion-based methods specifically designed for time series; CoRTX~\cite{chuang2023cortx} leveraging contrastive learning; SGT + GRAD~\cite{ismail2021improving} demonstrating an \textit{in-hoc} approach; and \textsc{TimeX}~\cite{queen2023encoding}, which serves as the strongest information-theoretic baseline. For \textbf{counterfactual explanations}, we compare against five advanced methodologies: CoMTE~\cite{COMTE}, AB-CF~\cite{li2023attention}, M-CELS~\cite{MCELS}, and CONFETTI~\cite{CONFETTI}. These baselines encapsulate a diverse range of optimization-based, evolutionary, saliency-guided, and multi-objective paradigms, ensuring a robust evaluation of counterfactual generation capabilities. 

\subsubsection{Evaluation Metrics}
To assess explanation quality across distinct paradigms, we use specific quantitative metrics. 

\textbf{Metrics for Attribution Explanations.}
Given that the precise salient features are known for the synthetic datasets, we utilize them as the ground truth for evaluation. At each time step, features causing prediction label changes are attributed an explanation of $1$, whereas those that do not affect such changes are $0$. Following the previous paper~\cite{crabbe2021explaining}, we evaluate the quality of salient features using Area Under Precision (AUP) and Area Under Recall (AUR), framing it as a binary classification task. 
We also employ AUPRC for consistency with TimeX~\cite{queen2023encoding}, which combines the results from both AUP and AUR. Higher values indicate better performance. Let $M \in [0,1]^{T \times D}$ be the obtained explanation mask, and $Q \in \{0,1\}^{T \times D}$ be the ground-truth matrix where $Q[t,d] = 1$ if feature $X[t, d]$ is salient. Given a detection threshold $\tau \in (0,1)$, we convert $M$ into a binary estimator $\hat{Q}[t,d](\tau)$:
\begin{equation}
    \hat{Q}[t,d](\tau) = \left\{\begin{array}{ll}1 & \text{ if } M[t,d] \geq \tau \\0 & \text{ else.}\end{array}\right.
\end{equation}
Considering the truly salient index set $A = \{(t,d) \mid Q[t,d] = 1\}$ and the selected index set $\hat{A}(\tau) = \{(t,d) \mid \hat{Q}[t,d](\tau) = 1\}$, the precision (P) and recall (R) curves are defined as:
\begin{equation}
\text{P}(\tau) = \frac{\vert A \cap \hat{A}(\tau) \vert}{\vert \hat{A}(\tau) \vert}, \quad \text{R}(\tau) = \frac{\vert A \cap \hat{A}(\tau) \vert}{\vert A \vert}.
\end{equation}
AUP and AUR are derived via integration over all thresholds:
\begin{equation}
\text{AUP} = \int_{0}^1 \text{P}(\tau) d\tau, \quad \text{AUR} = \int_{0}^1 \text{R}(\tau) d\tau.
\end{equation}

\begin{table*}[t]
    \centering
    \setlength{\tabcolsep}{5pt}
    \caption{Attribution explanation performance on univariate and multivariate datasets.}
    \resizebox{2.0\columnwidth}{!}{
    \begin{sc}
    \begin{tabular}{c|ccc|ccc|ccc }
         \toprule
         \textbf{Method} & \textbf{AUPRC} & \textbf{AUP} & \textbf{AUR} & \textbf{AUPRC} & \textbf{AUP} & \textbf{AUR} & \textbf{AUPRC} & \textbf{AUP} & \textbf{AUR}\\
         \midrule
         & \multicolumn{3}{c|}{\textbf{FreqShapes}} & \multicolumn{3}{c|}{\textbf{SeqComb-UV}} & \multicolumn{3}{c}{\textbf{ECG}}\\
         \rowcolor{gray!15}  IG & {0.752}\std{0.003} & {0.691}\std{0.003} & {0.598}\std{0.002} & {0.576}\std{0.002} & 0.816\std{0.002} & {0.287}\std{0.002} & {0.418}\std{0.001} & \underline{0.595}\std{0.002} & 0.320\std{0.001}\\
         Dynamask & 0.220\std{0.001} & 0.295\std{0.004} & 0.504\std{0.002} & 0.442\std{0.002} & {0.878}\std{0.004} & 0.103\std{0.001} & 0.328\std{0.001} & 0.525\std{0.003} & 0.108\std{0.008} \\
         \rowcolor{gray!15}  WinIT & 0.507\std{0.002} & 0.555\std{0.003} & 0.456\std{0.002} & 0.457\std{0.002} & 0.787\std{0.003} & 0.225\std{0.002} & 0.305\std{0.001} & 0.443\std{0.003} & {0.347}\std{0.001} \\
         CoRTX & 0.698\std{0.016} & 0.494\std{0.000} & 0.326\std{0.001} & 0.564\std{0.002} & 0.824\std{0.003} & 0.175\std{0.001} & 0.374\std{0.001} & 0.497\std{0.002} & 0.303\std{0.001}\\
         \rowcolor{gray!15}  SGT + Grad & 0.531\std{0.002} & 0.414\std{0.001} & 0.393\std{0.002} & 0.573\std{0.002} & 0.783\std{0.001} & 0.214\std{0.001} & 0.314\std{0.001} & 0.424\std{0.002} & 0.264\std{0.001} \\
         TimeX & \underline{0.832}\std{0.003} & \underline{0.722}\std{0.003} & \underline{0.638}\std{0.002} & \underline{0.712}\std{0.002} & \textbf{0.941}\std{0.001} & \underline{0.338}\std{0.001} & \underline{0.472}\std{0.002} & {0.566}\std{0.003} & \underline{0.446}\std{0.002} \\
          \midrule
         \rowcolor{mygreen} {\oursa} & \textbf{0.891}\std{0.002} & \textbf{0.781}\std{0.001} & \textbf{0.662}\std{0.002} & \textbf{0.847}\std{0.001} & \underline{0.907}\std{0.000} & \textbf{0.406}\std{0.001} & \textbf{0.660}\std{0.001} & \textbf{0.726}\std{0.001} & \textbf{0.460}\std{0.001} \\
         \midrule
         \midrule
         & \multicolumn{3}{c|}{\textbf{SeqComb-MV}} & \multicolumn{3}{c|}{\textbf{LowVar}} \\
         \rowcolor{gray!15}  IG & 0.330\std{0.002} & {0.748}\std{0.003} & 0.258\std{0.003} & \underline{0.869}\std{0.004} & {0.483}\std{0.003} & {0.817}\std{0.002}\\
         Dynamask & 0.314\std{0.002} & 0.548\std{0.005} & 0.195\std{0.003} & 0.139\std{0.001} & 0.164\std{0.003} & 0.211\std{0.002}\\
         \rowcolor{gray!15}  WinIT & 0.281\std{0.002} & 0.759\std{0.002} & 0.208\std{0.002} & 0.167\std{0.002} & 0.114\std{0.002} & 0.384\std{0.002} \\
         CoRTX & 0.363\std{0.002} & 0.563\std{0.001} & 0.346\std{0.002} & 0.498\std{0.001} & 0.328\std{0.003} & 0.471\std{0.001}\\
         \rowcolor{gray!15}  SGT + Grad & 0.489\std{0.001} & 0.497\std{0.001} & \textbf{0.429}\std{0.002} & 0.345\std{0.001} & 0.213\std{0.003} & 0.353\std{0.002}\\
         TimeX & \underline{0.688}\std{0.002} & \underline{0.833}\std{0.001} & {0.387}\std{0.002} & {0.867}\std{0.003} & \underline{0.545}\std{0.003} & \textbf{0.900}\std{0.002}\\
          \cmidrule{1-7}
         \rowcolor{mygreen}  {\oursa} & \textbf{0.759}\std{0.001} & \textbf{0.878}\std{0.001} & \underline{0.391}\std{0.001} & \textbf{0.947}\std{0.002} & \textbf{0.806}\std{0.002} & \underline{0.833}\std{0.002}\\
         \cmidrule[\heavyrulewidth]{1-7} 
    \end{tabular}
    \end{sc}
    }
    \label{tab:attr_synthetic}
\end{table*}

\begin{table*}[t]
\begin{center}
\setlength{\tabcolsep}{6pt}
\caption{Performance report on different real-world datasets by masking the top 10\% of salient features. The masked portion is substituted with an average of this feature or with zeros.}  
\label{top_feature}
\centering
\resizebox{1.7\columnwidth}{!}{
    \begin{sc}
\begin{tabular}{cc|ccccc|c}
\toprule
 \multicolumn{2}{c|}{\textbf{Method} \quad \textbf{Substitution}}& \textbf{PAM} & \textbf{Epilepsy} & \textbf{Boiler}& \textbf{Wafer} & \textbf{Freezer} & \textbf{Rank}\\
\midrule
 \rowcolor{gray!15} & Mean & 	0.979\std{0.002} & 0.939\std{0.005} & 0.814\std{0.056}& 0.993\std{0.002} & 0.772\std{0.093} & 7.2\\
 \rowcolor{gray!15} \multirow{-2}{*}{Random}  & Zero & 0.979\std{0.002} & 0.940\std{0.004} & 0.895\std{0.012}& 0.992\std{0.002} & 0.775\std{0.093} & 7.8\\
 
 & Mean & 0.778\std{0.015} & 0.816\std{0.019} & 0.808\std{0.036} & 0.488\std{0.267} & 0.452\std{0.129} & 4.6\\
 \multirow{-2}{*}{Dynamask}   & Zero & 	0.777\std{0.015} & \underline{0.361}\std{0.073} & 0.555\std{0.146} & 0.488\std{0.267} & \textbf{0.355}\std{0.110} & 3.6\\
 
 \rowcolor{gray!15} & Mean & \underline{0.749}\std{0.049} & 0.852\std{0.018} & 0.563\std{0.041} & 0.463\std{0.145} & 0.469\std{0.061} & 4.4\\
 \rowcolor{gray!15} \multirow{-2}{*}{TimeX}  & Zero & 0.754\std{0.049} & 0.454\std{0.077} & 0.416\std{0.048} & 0.465\std{0.057} & 0.465\std{0.057} & 3.6\\

\midrule

\rowcolor{mygreen} & Mean  & \textbf{0.717}\std{0.019} & 0.845\std{0.029} & \underline{0.385}\std{0.123} & \underline{0.410}\std{0.150} & 0.379\std{0.074}&\underline{2.8}\\
\rowcolor{mygreen} \multirow{-2}{*}{{\oursa}}  & Zero & 0.775\std{0.023} & \textbf{0.355}\std{0.146} & \textbf{0.361}\std{0.073} & \textbf{0.400}\std{0.061} &\underline{0.377}\std{0.074}&\textbf{2.0}\\
\bottomrule
\end{tabular}
    \end{sc}
} 
    \vspace{-3mm}
\end{center}
\end{table*}

\textbf{Metrics for Counterfactual Explanations.} {To comprehensively evaluate the quality of the generated counterfactual instances ($X''$), we assess them across five distinct dimensions: Validity, Confidence, Sparsity, Proximity-L1, and Proximity-L2. 
For Validity, Confidence, and Sparsity, higher values are better ($\uparrow$). For both Proximity metrics, lower values are better ($\downarrow$).
\begin{itemize}[leftmargin=*]
\item \textbf{Validity ($\uparrow$):} Measures the ratio of successful counterfactuals that successfully alter the black-box model's prediction to the desired target label $Y''$. For a dataset of $N$ instances:
\begin{equation}
\text{Validity} = \frac{1}{N} \sum_{i=1}^{N} \mathds{1}(f(X''_{i}) = Y'')
\end{equation}
\item \textbf{Confidence ($\uparrow$):} Evaluates the average predicted probability assigned by the classifier to the target class, indicating the robustness of the class flip:
\begin{equation}
\text{Confidence} = \frac{1}{N} \sum_{i=1}^{N} P_{f}(Y'' \mid X''_{i})
\end{equation}
\item \textbf{Sparsity ($\downarrow$):} Quantifies the minimality of the intervention by measuring the proportion of features left unperturbed. It is derived from the $L_1$ norm of the continuous mask $M$:
\begin{equation}
\text{Sparsity} = \frac{1}{N} \sum_{i=1}^{N} \left( \frac{1}{T \times D} \sum_{t=1}^{T} \sum_{d=1}^{D} M_i[t,d] \right)
\end{equation}
\item \textbf{Proximity-L1 ($\downarrow$):} Measures the feature-wise Mean Absolute Error (MAE) between the counterfactual explanation and the original input to ensure realistic generation:
\begin{equation}
\text{Proximity}_{L1} = \frac{1}{N} \sum_{i=1}^{N} \frac{1}{T \times D} ||X''_{i} - X_i||_1
\end{equation}
\item \textbf{Proximity-L2 ($\downarrow$):} Measures the feature-wise Mean Squared Error (MSE) to heavily penalize large, unrealistic deviations from the original input $X$:
\begin{equation}
\text{Proximity}_{L2} = \frac{1}{N} \sum_{i=1}^{N} \frac{1}{T \times D} ||X''_{i} - X_i||_2^2
\end{equation}
\end{itemize}
}

\subsection{ Efficacy of Unified Explanations}
\label{sec:exp_efficacy}
We first examine the fidelity and efficacy of \modelname~in producing both attribution and counterfactual explanations via the single unified training process. 

\begin{figure}[t]
    \centering
    \begin{subfigure}{0.31\linewidth}
        \centering
        \includegraphics[width=\linewidth]{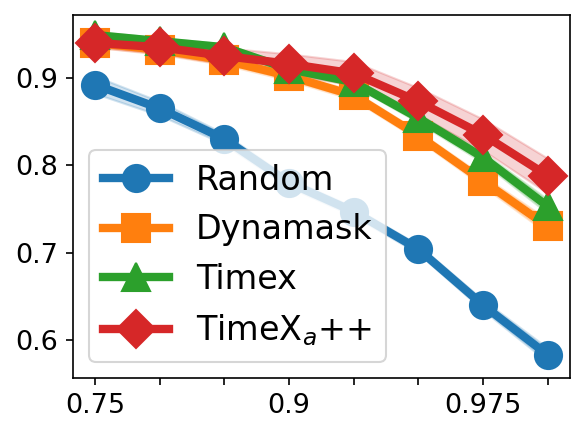}
        \caption{PAM}
    \end{subfigure}
    \begin{subfigure}{0.31\linewidth}
        \centering
        \includegraphics[width=\linewidth]{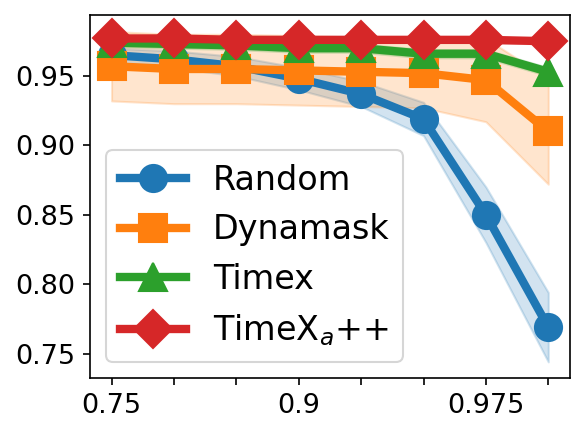}
        \caption{Epilepsy}
    \end{subfigure}
    \begin{subfigure}{0.31\linewidth}
        \centering
        \includegraphics[width=\linewidth]{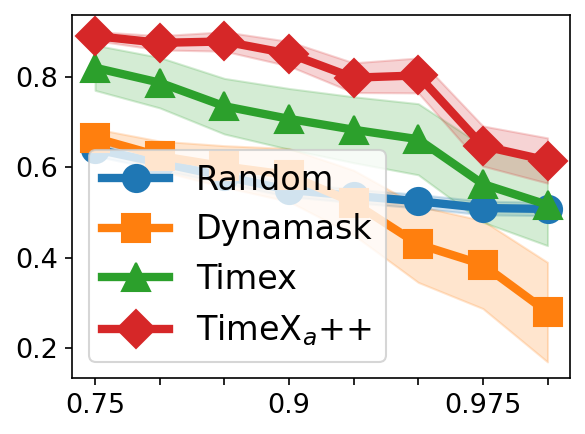}
        \caption{Boiler}
    \end{subfigure}
    \caption{Occlusion experiments on real-world datasets. Higher values indicate better performance. The x-axis is the Bottom Proportion Perturbed, and the y-axis is the Prediction AUROC.}
    \vspace{-4mm}
    \label{fig:occlusion}
\end{figure}
For \textbf{attribution} explanations, Table~\ref{tab:attr_synthetic} summarizes the performance across univariate and multivariate synthetic datasets. {\oursa}~outperforms competing explainers in $9$ out of $12$ evaluation cases. Compared to the strongest baseline, \textsc{TimeX}, our framework achieves average improvements of $11.01\%$ in AUPRC, $10.87\%$ in AUP, and $1.25\%$ in AUR. 
When analyzing the global metric AUPRC specifically, {\oursa}~significantly improves upon the identification of ground-truth explanations by $6.97\%$ on FreqShapes, $18.86\%$ on SeqComb-UV, $10.33\%$ on SeqComb-MV, and $8.91\%$ on LowVar over the strongest baselines. 
The statistical significance of these improvements is corroborated by the Friedman Test across all methods and cases, yielding a statistic of $F_F=51.32$ with $p<0.001$. It supports our theoretical claim that optimizing via the Information Bottleneck principle reliably extracts precise temporal explanation instances without compromising predictive accuracy.

\begin{table*}[t]
    \centering
    \setlength{\tabcolsep}{3pt}
    \caption{{Targeted counterfactual explanation performance on univariate and multivariate datasets.}}
    \resizebox{2.0\columnwidth}{!}{
    \begin{sc}
    \begin{tabular}{c|ccccc|ccccc}
         \toprule
         
         \textbf{Method} & \textbf{Validity} & \textbf{Confidence} & \textbf{Sparsity} & \textbf{Prox.-L1} & \textbf{Prox.-L2} & \textbf{Validity} & \textbf{Confidence} & \textbf{Sparsity}  & \textbf{Prox.-L1} & \textbf{Prox.-L2} \\
         \midrule
         & \multicolumn{5}{c|}{\textbf{FreqShapes}} & \multicolumn{5}{c}{\textbf{SeqComb-UV}} \\
         \rowcolor{gray!15} CoMTE &  1.000 \std{ 0.000	}&	0.965 \std{ 0.006	}&	1.000 \std{ 0.000	}&	0.573 \std{ 0.004	}&	0.828 \std{ 0.006	}&  1.000 \std{ 0.000	}&	0.784 \std{ 0.022	}&	0.999 \std{ 0.000	}&	0.209 \std{ 0.002	}&	0.401 \std{ 0.008	} \\
         AB-CF & 1.000 \std{ 0.000	}&	0.871 \std{ 0.008	}&	0.733 \std{ 0.025	}&	0.441 \std{ 0.013	}&	0.718 \std{ 0.011	}& 1.000 \std{ 0.000 	}&	0.911 \std{ 0.013	}&	0.650 \std{ 0.037	}&	0.151 \std{ 0.007	}&	0.348 \std{ 0.014	} \\
         \rowcolor{gray!15} M-CELS & 0.486 \std{ 0.053	}&	0.466 \std{ 0.051	}&	0.061 \std{ 0.008	}&	0.064 \std{ 0.011	}&	0.196 \std{ 0.031	}& 0.361 \std{ 0.037	}&	0.353 \std{ 0.036	}&	0.025 \std{ 0.003	}&	0.016 \std{ 0.001	}&	0.090 \std{ 0.007	} \\
         CONFETTI & 0.990 \std{ 0.003	}&	0.512 \std{ 0.002	}&	0.349 \std{ 0.010	}&	0.242 \std{ 0.011	}&	0.538 \std{ 0.012	}& 0.894 \std{ 0.034	}&	0.477 \std{ 0.017	}&	0.209 \std{ 0.019	}&	0.059 \std{ 0.008	}&	0.222 \std{ 0.021	}\\
          \midrule
         \rowcolor{mygreen} {\ourscf} & 0.974 \std{ 0.004	}&	0.964 \std{ 0.005	}&	0.214 \std{ 0.005	}&	0.339 \std{ 0.021	}&	0.757 \std{ 0.030	}& 0.890 \std{ 0.053	}&	0.887 \std{ 0.052	}&	0.516 \std{ 0.074	}&	0.288 \std{ 0.085	}&	0.487 \std{ 0.082	}\\
         \midrule
         \midrule
         & \multicolumn{5}{c|}{\textbf{SeqComb-MV}} & \multicolumn{5}{c}{\textbf{LowVar}} \\
         \rowcolor{gray!15}  CoMTE &  0.990 \std{ 0.004	}&	0.820 \std{ 0.020	}&	0.443 \std{ 0.018	}&	0.081 \std{ 0.003	}&	0.247 \std{ 0.006	}& 1.000 \std{ 0.000	}&	0.985 \std{ 0.003	}&	0.733 \std{ 0.007	}&	0.780 \std{ 0.007	}&	1.121 \std{ 0.005	} \\
         AB-CF &  1.000 \std{ 0.000	}&	0.954 \std{ 0.008	}&	0.641 \std{ 0.029	}&	0.099 \std{ 0.004	}&	0.246 \std{ 0.007	}& 1.000 \std{ 0.000 	}&	0.916 \std{ 0.009	}&	0.627 \std{ 0.011	}&	0.662 \std{ 0.012	}&	1.003 \std{ 0.011	} \\
         \rowcolor{gray!15} M-CELS & 0.350 \std{ 0.043	}&	0.348 \std{ 0.042	}&	0.011 \std{ 0.001	}&	0.008 \std{ 0.001	}&	0.072 \std{ 0.007	}& 0.365 \std{ 0.017	}&	0.357 \std{ 0.015	}&	0.029 \std{ 0.001	}&	0.039 \std{ 0.001	}&	0.232 \std{ 0.005	} \\
         CONFETTI & 0.910 \std{ 0.031	}&	0.475 \std{ 0.013	}&	0.246 \std{ 0.017	}&	0.039 \std{ 0.003	}&	0.153 \std{ 0.011	}& 0.998 \std{ 0.001	}&	0.514 \std{ 0.000	}&	0.190 \std{ 0.004	}&	0.193 \std{ 0.005	}&	0.537 \std{ 0.007	} \\
          \midrule
         \rowcolor{mygreen} {\ourscf} & 0.983 \std{ 0.007	}&	0.979 \std{ 0.008	}&	0.305 \std{ 0.041	}&	0.152 \std{ 0.016	}&	0.353 \std{ 0.033	}& 0.985 \std{ 0.003	}&	0.980 \std{ 0.003	}&	0.175 \std{ 0.024	}&	0.212 \std{ 0.010	}&	0.548 \std{ 0.013	} \\
         \midrule
         \midrule
         & \multicolumn{5}{c|}{\textbf{ECG}} & \multicolumn{5}{c}{\textbf{Epilepsy}} \\
         \rowcolor{gray!15}  CoMTE &  1.000 \std{ 0.000	}&	0.936 \std{ 0.008	}&	1.000 \std{ 0.000	}&	0.275 \std{ 0.006	}&	0.380 \std{ 0.010	}& 1.000 \std{ 0.000	}&	0.852 \std{ 0.027	}&	0.996 \std{ 0.000	}&	0.633 \std{ 0.130	}&	0.832 \std{ 0.162	} \\
         AB-CF & 1.000 \std{ 0.000	}&	0.841 \std{ 0.008	}&	0.701 \std{ 0.059	}&	0.189 \std{ 0.017	}&	0.303 \std{ 0.018	}& 1.000 \std{ 0.000	}&	0.739 \std{ 0.036	}&	0.788 \std{ 0.039	}&	0.539 \std{ 0.132	}&	0.762 \std{ 0.161	} \\
        \rowcolor{gray!15}  M-CELS & 0.231 \std{ 0.075	}&	0.242 \std{ 0.068	}&	0.022 \std{ 0.005	}&	0.013 \std{ 0.003	}&	0.061 \std{ 0.011	}& 0.097 \std{ 0.049	}&	0.103 \std{ 0.046	}&	0.007 \std{ 0.005	}&	0.009 \std{ 0.006	}&	0.027 \std{ 0.017	} \\
         CONFETTI & 0.977 \std{ 0.005	}&	0.505 \std{ 0.003	}&	0.287 \std{ 0.008	}&	0.080 \std{ 0.003	}&	0.200 \std{ 0.007	}& 0.798 \std{ 0.050	}&	0.423 \std{ 0.026	}&	0.334 \std{ 0.026	}&	0.212 \std{ 0.018	}&	0.417 \std{ 0.030	}\\
          \midrule
         \rowcolor{mygreen} {\ourscf} & 0.848 \std{ 0.104	}&	0.847 \std{ 0.101	}&	0.084 \std{ 0.026	}&	0.251 \std{ 0.097	}&	0.767 \std{ 0.185	}& 0.946 \std{ 0.047	}&	0.924 \std{ 0.045	}&	0.139 \std{ 0.024	}&	0.255 \std{ 0.028	}&	0.698 \std{ 0.041	}\\
         \bottomrule
    \end{tabular}
    \end{sc}
    }
    \label{tab:cf_target}
\end{table*}

\begin{table*}[t]
    \centering
    \setlength{\tabcolsep}{3pt}
    \caption{{Untargeted counterfactual explanation performance on univariate and multivariate datasets.}}
    \resizebox{2.0\columnwidth}{!}{
    \begin{sc}
    \begin{tabular}{c|ccccc|ccccc}
         \toprule
         \textbf{Method} & \textbf{Validity} & \textbf{Confidence} & \textbf{Sparsity}  & \textbf{Prox.-L1} & \textbf{Prox.-L2} & \textbf{Validity} & \textbf{Confidence} & \textbf{Sparsity}  & \textbf{Prox.-L1} & \textbf{Prox.-L2} \\
         \midrule
         & \multicolumn{5}{c|}{\textbf{FreqShapes}} & \multicolumn{5}{c}{\textbf{SeqComb-UV}} \\
         \rowcolor{gray!15} CoMTE & 1.000 \std{ 0.000	}&	0.930 \std{ 0.017	}&	1.000 \std{ 0.000	}&	0.438 \std{ 0.016	}&	0.641 \std{ 0.028	}& 1.000 \std{ 0.000	}&	0.790 \std{ 0.023	}&	0.999 \std{ 0.000	}&	0.196 \std{ 0.000	}&	0.376 \std{ 0.001	} \\
         AB-CF & 1.000 \std{ 0.000	}&	0.826 \std{ 0.021	}&	0.672 \std{ 0.006	}&	0.283 \std{ 0.010	}&	0.486 \std{ 0.021	}& 1.000 \std{ 0.000	}&	0.890 \std{ 0.018	}&	 0.513 \std{ 0.040	}&	0.114 \std{ 0.004	}&	0.293 \std{ 0.004	} \\
         \rowcolor{gray!15} M-CELS & 0.612 \std{ 0.039	}&	0.570 \std{ 0.034	}&	0.063 \std{ 0.006	}&	0.040 \std{ 0.001	}&	0.157 \std{ 0.002	}& 0.336 \std{ 0.026	}&	0.319 \std{ 0.022	}&	0.013 \std{ 0.001	}&	0.011 \std{ 0.001	}&	0.083 \std{ 0.004	} \\
         CONFETTI & 0.621 \std{ 0.017	}&	0.514 \std{ 0.019	}&	0.257 \std{ 0.008	}&	0.117 \std{ 0.007	}&	0.260 \std{ 0.017	}& 0.962 \std{ 0.010	}&	0.519 \std{ 0.008	}&	0.200 \std{ 0.005	}&	0.052 \std{ 0.002	}&	0.208 \std{ 0.005	} \\
          \midrule
         \rowcolor{mygreen} {\ourscf} &1.000 \std{ 0.000	}&	0.999 \std{ 0.000	}&	0.233 \std{ 0.009	}&	0.420 \std{ 0.016	}&	0.872 \std{ 0.020	}&  1.000 \std{ 0.000	}&	0.999 \std{ 0.001	}&	0.624 \std{ 0.082	}&	0.393 \std{ 0.069	}&	0.637 \std{ 0.076	} \\
         \midrule
         \midrule
         & \multicolumn{5}{c|}{\textbf{SeqComb-MV}} & \multicolumn{5}{c}{\textbf{LowVar}} \\
         \rowcolor{gray!15} CoMTE & 1.000 \std{ 0.000	}&	0.796 \std{ 0.022	}&	0.367 \std{ 0.013	}&	0.063 \std{ 0.001	}&	0.208 \std{ 0.002	}& 1.000 \std{ 0.000	}&	0.978 \std{ 0.003	}&	0.520 \std{ 0.002	}&	0.552 \std{ 0.002	}&	0.951 \std{ 0.002	} \\
         AB-CF & 1.000 \std{ 0.000	}&	0.941 \std{ 0.013	}&	0.514 \std{ 0.018	}&	0.075 \std{ 0.002	}&	0.202 \std{ 0.003	}& 1.000 \std{ 0.000	}&	0.866 \std{ 0.014	}&	0.556 \std{ 0.020	}&	0.579 \std{ 0.020	}&	0.916 \std{ 0.019	}\\
         \rowcolor{gray!15} M-CELS & 0.158 \std{ 0.026	}&	0.159 \std{ 0.026	}&	0.005 \std{ 0.000	}&	0.004 \std{ 0.000	}&	0.051 \std{ 0.005	}& 0.312 \std{ 0.045	}&	0.296 \std{ 0.037	}&	0.021 \std{ 0.001	}&	0.030 \std{ 0.001	}&	0.206 \std{ 0.006	} \\
         CONFETTI & 0.917 \std{ 0.037	}&	0.483 \std{ 0.014	}&	0.270 \std{ 0.010	}&	0.044 \std{ 0.001	}&	0.169 \std{ 0.004	}& 0.998 \std{ 0.001	}&	0.513 \std{ 0.001	}&	0.176 \std{ 0.004	}&	0.175 \std{ 0.004	}&	0.506 \std{ 0.006	} \\
          \midrule
         \rowcolor{mygreen} {\ourscf} & 1.000 \std{ 0.000	}&	1.000 \std{ 0.000	}&	0.378 \std{ 0.092	}&	0.118 \std{ 0.010	}&	0.291 \std{ 0.008	}& 1.000 \std{ 0.000	}&	0.999 \std{ 0.000	}&	0.221 \std{ 0.049	}&	0.231 \std{ 0.014	}&	0.552 \std{ 0.023	}\\
         \bottomrule
    \end{tabular}
    \end{sc}
    }
    \vspace{-2mm}
    \label{tab:cf_untarget}
\end{table*}

Furthermore, to evaluate attribution fidelity on real-world datasets where ground-truth explanations are unavailable, we conduct occlusion experiments. Specifically, we occlude the bottom $k$-percentile of salient features to measure the change in prediction AUROC. Besides the standard baselines, we also include a random explainer reference to control for potential misinterpretations. The explanation results by occluding salient features are presented in Figure~\ref{fig:occlusion}. 
Our results show that {\oursa}~outperforms others across both univariate (Epilepsy) and multivariate (PAM and Boiler) time series. Particularly, {\oursa}~maintains non-decreasing performance on Epilepsy due to the retention of only salient features, and outperforms baselines at any threshold $k$ on PAM and Boiler datasets. Moreover, our method maintains excellent stability compared to the strongest baseline \textsc{TimeX}, where the error bars of our method are noticeably narrower. This is because {\oursa}~avoids the label leakage caused by re-optimizing a white-box predictive model. We also delete the top 10\% of salient features and substitute them with an average of this feature or with zero perturbations~\citep{liu2024explaining}, and further expand our experiments to five real-world time series datasets. The results (Table~\ref{top_feature}) consistently show that the proposed method outperforms existing explainers under different perturbations, with substitution with zeros being the most applicable and yielding the highest average rank. The statistical significance of these improvements is firmly corroborated by the Friedman Test yielding $F_F=25.80$ and $p<0.001$.

Transitioning to \textbf{counterfactual} reasoning, we compare {\ourscf}~with state-of-the-art baselines across six benchmark datasets under a targeted setting, rigorously evaluating all possible target labels for each prediction, as presented in Table~\ref{tab:cf_target}. Given the multi-objective nature of counterfactual explanations, directly comparing individual evaluation metrics is insufficient.  Ideally, counterfactual explanations should exhibit low sparsity, low proximity, high validity, and high confidence. However, these objectives are often conflicting, resulting in an inherent trade-off among sparsity, validity, and proximity. Therefore, we adopt Pareto frontier analysis to demonstrate their overall effectiveness. 
Among the baselines, CONFETTI achieves high validity scores; however, its relatively low confidence indicates that the generated counterfactual explanations tend to lie close to the decision boundary. M-CELS generally produces counterfactual explanations with low sparsity but underperforms in terms of validity and confidence. 
AB-CF and COMTE achieve high validity and confidence, but they exhibit substantially higher sparsity, requiring extensive modifications to the input, which may generate target-labeled references.
In contrast, {\ourscf}~achieves a more favorable balance between validity/confidence and sparsity/proximity. Across these datasets, our method attains competitive validity and confidence compared with AB-CF and COMTE while maintaining substantially lower sparsity and proximity. For instance, on the Epilepsy dataset, {\ourscf}~achieves the highest confidence score while requiring only 0.14 average sparsity, demonstrating the effectiveness of the proposed framework.

\begin{table*}[t]
    \centering
    \caption{Difference between the distribution of different explanation instances and the distribution of original data.}
    \resizebox{2.0\columnwidth}{!}{
    \begin{sc}
    \begin{tabular}{c|ccc|ccc}
         \toprule
         \textbf{Method} & \textbf{KDE} $\uparrow$& \textbf{KL-Divergence} $\downarrow $& \textbf{MMD} $\downarrow $& \textbf{KDE} $\uparrow$& \textbf{KL-Divergence} $\downarrow $& \textbf{MMD} $\downarrow $\\
         \midrule
         & \multicolumn{3}{c|}{\textbf{FreqShapes}} & \multicolumn{3}{c}{\textbf{SeqComb-UV}} \\
         \rowcolor{gray!15} Zero& -36.671\std{ 0.275} & \underline{0.144}\std{0.007} & 0.077\std{0.004} & -90.093\std{ 0.312} & 0.399\std{0.005} & 0.167\std{0.003} \\
         Mean &\underline{-36.539}\std{0.194} & 0.147\std{0.003} & 0.078\std{0.006} &\underline{-83.038}\std{0.271} & \underline{0.325}\std{0.009} & 0.111\std{0.005}\\
         \rowcolor{gray!15} $b\sim \mathbb{B}_\mathcal{X}$ & -53.855\std{0.509} & 0.289\std{0.003} & \underline{0.024}\std{0.000} &-100.050\std{0.658} & 0.498\std{0.005} & \textbf{0.009}\std{0.000}\\
         \midrule
         \rowcolor{mygreen}  {\oursa} & \textbf{-28.757}\std{2.658} & \textbf{0.096}\std{0.024} & \textbf{0.016}\std{0.004} & \textbf{-57.932}\std{1.684} & \textbf{0.060}\std{0.018} & \underline{0.078}\std{0.019}\\
         \midrule
         \midrule
         & \multicolumn{3}{c|}{\textbf{SeqComb-MV}} & \multicolumn{3}{c}{\textbf{LowVar}} \\
         \rowcolor{gray!15} Zero & -257.040\std{1.656} & 0.739\std{0.029} & 0.248\std{0.006}& -431.993\std{1.131} & 0.597\std{0.012} & 0.105\std{0.003}\\
         Mean &\underline{-230.250}\std{1.313} & \underline{0.455}\std{0.019} & 0.074\std{0.009} &\underline{-429.331}\std{1.249} & \underline{0.571}\std{0.012} & \underline{0.095}\std{0.004}\\
         \rowcolor{gray!15} $b\sim \mathbb{B}_\mathcal{X}$ &-260.182\std{1.571} & 0.721\std{0.019} & \underline{0.026}\std{0.000} &-474.482\std{1.319} & 0.976\std{0.017} & 0.104\std{0.000}\\
         \midrule
         \rowcolor{mygreen}  {\oursa} & \textbf{-191.265}\std{0.890} & \textbf{0.038}\std{0.010} & \textbf{0.014}\std{0.012} & \textbf{-426.431}\std{2.765} & \textbf{0.538}\std{0.025} & \textbf{0.073}\std{0.017}\\
\bottomrule
    \end{tabular}
    \end{sc}
    }
    \vspace{-1mm}
    \label{tab:perturbations}
\end{table*}

\begin{table*}[t]
  \centering
\setlength{\tabcolsep}{3pt}
  \caption{Explainer results with LSTM and CNN predictors on FreqShapes and SeqComb-MV synthetic datasets.}
  \vspace{1mm}
      \resizebox{2.0\columnwidth}{!}{
    \begin{sc}
  \begin{tabular}{c|ccc|ccc|ccc}
    \toprule
     & \multicolumn{3}{c|}{\textbf{FreqShapes}} & \multicolumn{3}{c|}{\textbf{SeqComb-MV}}& \multicolumn{3}{c}{\textbf{ECG}}\\
     \textbf{Method} & \textbf{AUPRC} & \textbf{AUP} & \textbf{AUR} & \textbf{AUPRC} & \textbf{AUP} & \textbf{AUR}  & \textbf{AUPRC} & \textbf{AUP} & \textbf{AUR} \\
    \midrule
     \multicolumn{10}{c}{\textbf{LSTM}}\\
     \rowcolor{gray!15} IG & 0.928\std{0.002} & 0.778\std{0.001} & 0.693\std{0.002} & 0.237\std{0.002} & 0.515\std{0.005} & 0.321\std{0.003} & 0.504\std{0.002} & 0.613\std{0.003} & 0.403\std{0.002} \\
     Dynamask & 0.229\std{0.001} & 0.342\std{0.004} & 0.517\std{0.001} & 0.284\std{0.002} & 0.637\std{0.005} & 0.182\std{0.002}  & 0.373\std{0.001} & 0.630\std{0.003} & 0.110\std{0.001} \\
     \rowcolor{gray!15} WinIT & 0.417\std{0.002} & 0.511\std{0.003} & 0.391\std{0.002} & \underline{0.352}\std{0.001} & \underline{0.655}\std{0.003} & 0.342\std{0.002}  & 0.363\std{0.001} & 0.381\std{0.002} & 0.406\std{0.001} \\
     TimeX & \underline{0.990}\std{0.000} & \textbf{0.789}\std{0.001} & \underline{0.796}\std{0.001} & 0.130\std{0.002} & 0.131\std{0.002} & \textbf{0.475}\std{0.002} & \underline{0.606}\std{0.002} & \underline{0.642}\std{0.002} & \underline{0.444}\std{0.002} \\
     \midrule
     \rowcolor{mygreen} {\oursa} & \textbf{0.994}\std{0.000} & \underline{0.741}\std{0.001} & \textbf{0.843}\std{0.001} & \textbf{0.405}\std{0.004} & \textbf{0.680}\std{0.005} & \underline{0.352}\std{0.002} & \textbf{0.651}\std{0.001} & \textbf{0.743}\std{0.001} & \textbf{0.445}\std{0.001} \\
    \hline
    \hline
     \multicolumn{10}{c}{\textbf{CNN}}\\
     \rowcolor{gray!15} IG & \textbf{0.991}\std{0.001} & \textbf{0.878}\std{0.001} & 0.706\std{0.002} & 0.598\std{0.003} & \underline{0.886}\std{0.001} & 0.229\std{0.001} & 0.495\std{0.001} & 0.537\std{0.001} & \textbf{0.531}\std{0.001}\\
     Dynamask & 0.257\std{0.001} & 0.443\std{0.003} & 0.526\std{0.002} & 0.455\std{0.002} & 0.731\std{0.003} & 0.314\std{0.002}& 0.460\std{0.001} & 0.722\std{0.003} & 0.131\std{0.001} \\
     \rowcolor{gray!15} WinIT &  0.532\std{0.002} & 0.602\std{0.003} & 0.397\std{0.002} & 0.533\std{0.001} & 0.832\std{0.002} & 0.226\std{0.002} & 0.396\std{0.001} & 0.329\std{0.002} & 0.352\std{0.001} \\
     TimeX & 0.749\std{0.005} & 0.497\std{0.003} & \underline{0.792}\std{0.002} & \underline{0.702}\std{0.002} & 0.767\std{0.001} & \textbf{0.469}\std{0.002} & \underline{0.640}\std{0.001} & \underline{0.746}\std{0.001} & 0.416\std{0.001} \\
     \midrule
     \rowcolor{mygreen} {\oursa} & \underline{0.913}\std{0.001} & \underline{0.607}\std{0.001} & \textbf{0.795}\std{0.001} & \textbf{0.782}\std{0.001} & \textbf{0.890}\std{0.001} & \underline{0.343}\std{0.001} & \textbf{0.673}\std{0.001} & \textbf{0.757}\std{0.001} & \underline{0.432}\std{0.001}\\
    \bottomrule
  \end{tabular}
    \end{sc}
  }
  \label{table:lstm_synth}
\end{table*}

To further validate the effectiveness of {\ourscf}, we conduct experiments under the untargeted setting using four synthetic datasets. For implementation, multiple target labels are considered, and the counterfactual explanation with the highest confidence is selected as the final result. As shown in Table~\ref{tab:cf_untarget}, {\ourscf}~demonstrates performance trends consistent with those observed in the targeted setting. Baseline methods that rely heavily on subsequence search, such as CONFETTI, exhibit limited robustness to the selection of reference data. In contrast, {\ourscf}~consistently achieves the highest validity and confidence across all four datasets, highlighting its superior robustness compared with the baselines.

\subsection{Resilience to Out-of-Distribution Shift}
\label{sec:exp_ood}
A critical vulnerability of perturbation-based attribution methods is the OOD shift, where explanation instances fall outside the original data manifold, yielding unreliable model gradients. We systematically quantify this phenomenon using Kernel Density Estimation (KDE)~\cite{parzen1962estimation}, KL divergence~\cite{kullback1951information}, and Maximum Mean Discrepancy (MMD)~\cite{gretton2012kernel} to assess the distributional alignment between the explanation-embedded instances and the original temporal sequences. 
Specifically, KDE provides a non-parametric estimation of the probability density function of the original data manifold; by evaluating the log-likelihood of perturbed instances under this density, KDE effectively identifies OOD samples, as instances falling into low-density regions exhibit significantly lower likelihood scores~\cite{parzen1962estimation, roth2022out}. 
Furthermore, MMD serves as a robust kernel-based two-sample test that measures the distance between the mean embeddings of two distributions within a Reproducing Kernel Hilbert Space (RKHS). A smaller MMD value formally indicates that the distribution of the generated explanation instances is statistically indistinguishable from the original empirical distribution, directly quantifying the mitigation of the OOD shift~\cite{gretton2012kernel, rabanser2019failing}. 
Table~\ref{tab:perturbations} reveals that explanations generated by {\oursa}~exhibit significantly lower MMD and KL divergence compared to conventional masking strategies.

\subsection{Robustness Against Noise}
\label{sec:exp_adv}
Beyond OOD issues in attribution, counterfactual explanations frequently suffer from adversarial degeneration, where imperceptible, high-frequency noise flips the model's prediction without providing meaningful semantic transitions. We demonstrate that {\ourscf}~eradicates this vulnerability by strictly confining perturbations. The proximity metrics analyzed alongside sparsity indicate that our framework produces highly localized structural changes rather than distributed trivial noise.
This structural integrity is governed by the causal structural anchor $\mathcal{L}_{bound}$, acting as a stringent penalty against any perturbation occurring in the background. By forcing the generator $\psi_{cf}$ to manipulate only the critical temporal windows identified by the bottleneck extractor, the resulting counterfactuals are guaranteed to represent semantic shifts.

\begin{table}[t]
\begin{center}
\setlength{\tabcolsep}{2pt}
\caption{Inference runtime of occlusion experiments for IG, Dynamask, \textsc{TimeX}, and {\oursa}~on all real-world datasets.}\label{times}
\centering
\vspace{1mm}
\resizebox{1\columnwidth}{!}{
    \begin{sc}
\begin{tabular}{c|ccccc}
\toprule
 \textbf{Method} & \textbf{PAM} & \textbf{Epilepsy} & \textbf{Boiler} \\
\midrule
 \rowcolor{gray!15} IG & 5.231\std{0.247} & 15.467\std{0.363} & 123.556\std{0.807} \\
 Dynamask & 105.166\std{0.610} & 371.092\std{4.257} & 3780.433\std{468.144} \\
  \rowcolor{gray!15} TimeX & \underline{0.164}\std{0.219} & \underline{0.204}\std{0.207} & \underline{0.918}\std{0.203}\\
 \midrule
\rowcolor{mygreen} {\oursa} & \textbf{0.161}\std{0.218} & \textbf{0.201}\std{0.208} & \textbf{0.910}\std{0.177}\\
 \midrule
 \midrule
 \textbf{Method} & \textbf{Wafer} & \textbf{FreezerRegular} \\
 \cmidrule{1-3}
  \rowcolor{gray!15} IG & 15.395\std{0.305} & 6.745\std{0.502} \\
 Dynamask & 403.400\std{31.664} & 189.285\std{1.731}\\
  \rowcolor{gray!15} TimeX & \underline{0.163}\std{0.022} & \textbf{0.269}\std{0.411}\\
 \cmidrule{1-3}
\rowcolor{mygreen} {\oursa}  & \textbf{0.126}\std{0.002} & \underline{0.270}\std{0.417}\\
\cmidrule[\heavyrulewidth]{1-3}
\end{tabular}
    \end{sc}
} 
\end{center}
\vspace{-4mm}
\end{table}

\begin{table}[t]
\begin{center}
\setlength{\tabcolsep}{3pt}
\caption{{Inference runtime of \textbf{counterfactual} explanation experiments for one split.}}
\label{tab:runtime_cf}
\centering
\resizebox{1\columnwidth}{!}{
    \begin{sc}
\begin{tabular}{c|ccc}
\toprule
 \textbf{Method} & \textbf{FreqShapes} & \textbf{LowVar} & \textbf{Epilepsy}   \\
\midrule
  \rowcolor{gray!15} CoMTE     & 42.245\std{ 1.146	}   & 64.708 \std{ 8.603	}    &	22.992 \std{ 0.858	} \\
 AB-CF     & 39.042\std{ 1.139	}   & 37.944\std{2.479}        &	32.750 \std{ 2.522	} \\
  \rowcolor{gray!15} M-CELS    & 4271.181\std{55.677}  & 5411.136\std{63.176}     &  1887.034\std{69.215}    \\
 CONFETTI  & 6628.330\std{103.100} & 8397.309\std{172.012}    &  4519.880\std{453.902}    \\
 \midrule
  \rowcolor{mygreen} {\ourscf} & \textbf{0.075} \std{ \textbf{0.035}}    & \textbf{0.287} \std{ \textbf{0.119}}    &  \textbf{0.140}\std{\textbf{0.012}}   \\
\bottomrule
\end{tabular}
    \end{sc}
} 
\end{center}
\vspace{-4mm}
\end{table}

\begin{figure}[t]	
	\centering
    \begin{subfigure}{0.45\columnwidth}
        \centering
        \includegraphics[width=\linewidth]{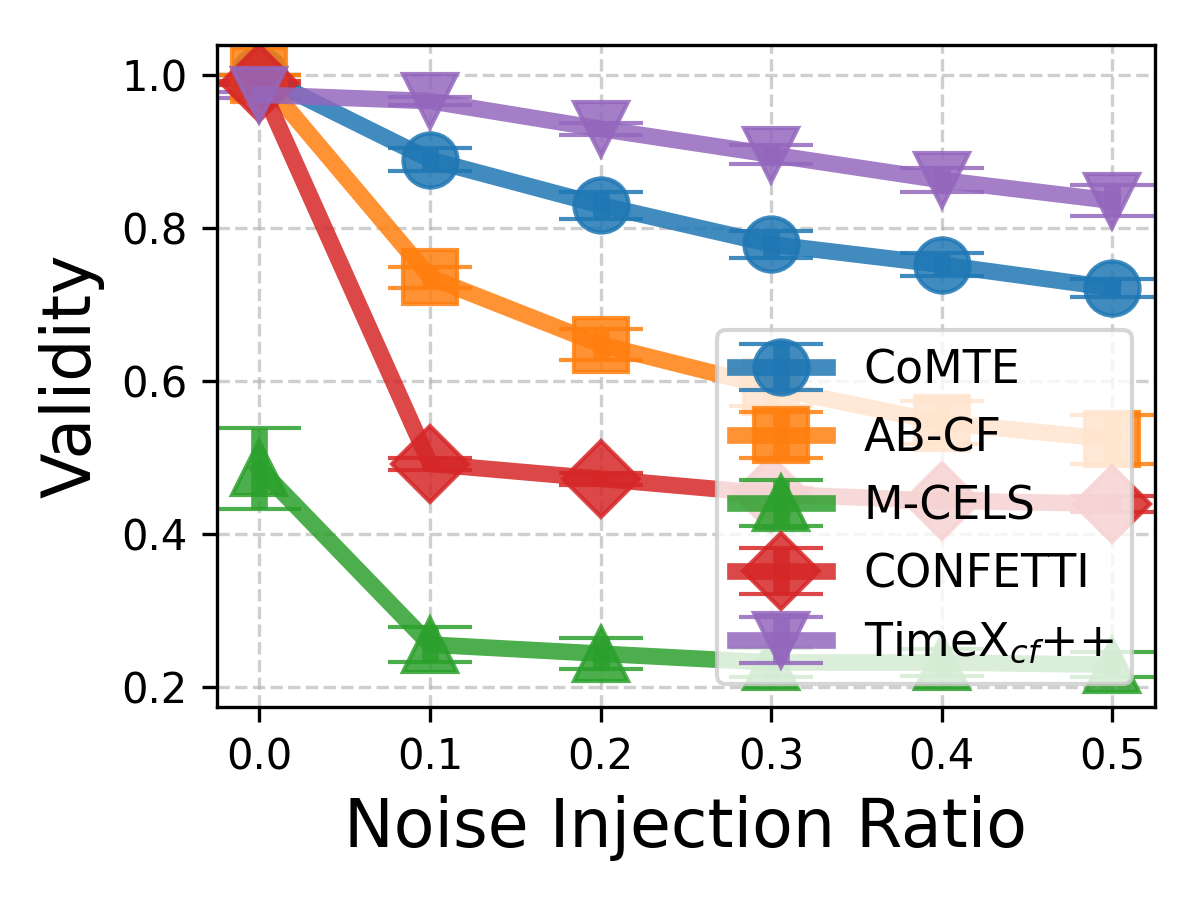}
        \caption{Validity on Freqshape}
        \label{cf_rob_fig:val_freq}
    \end{subfigure}
    \begin{subfigure}{0.45\columnwidth}
        \centering
        \includegraphics[width=\linewidth]{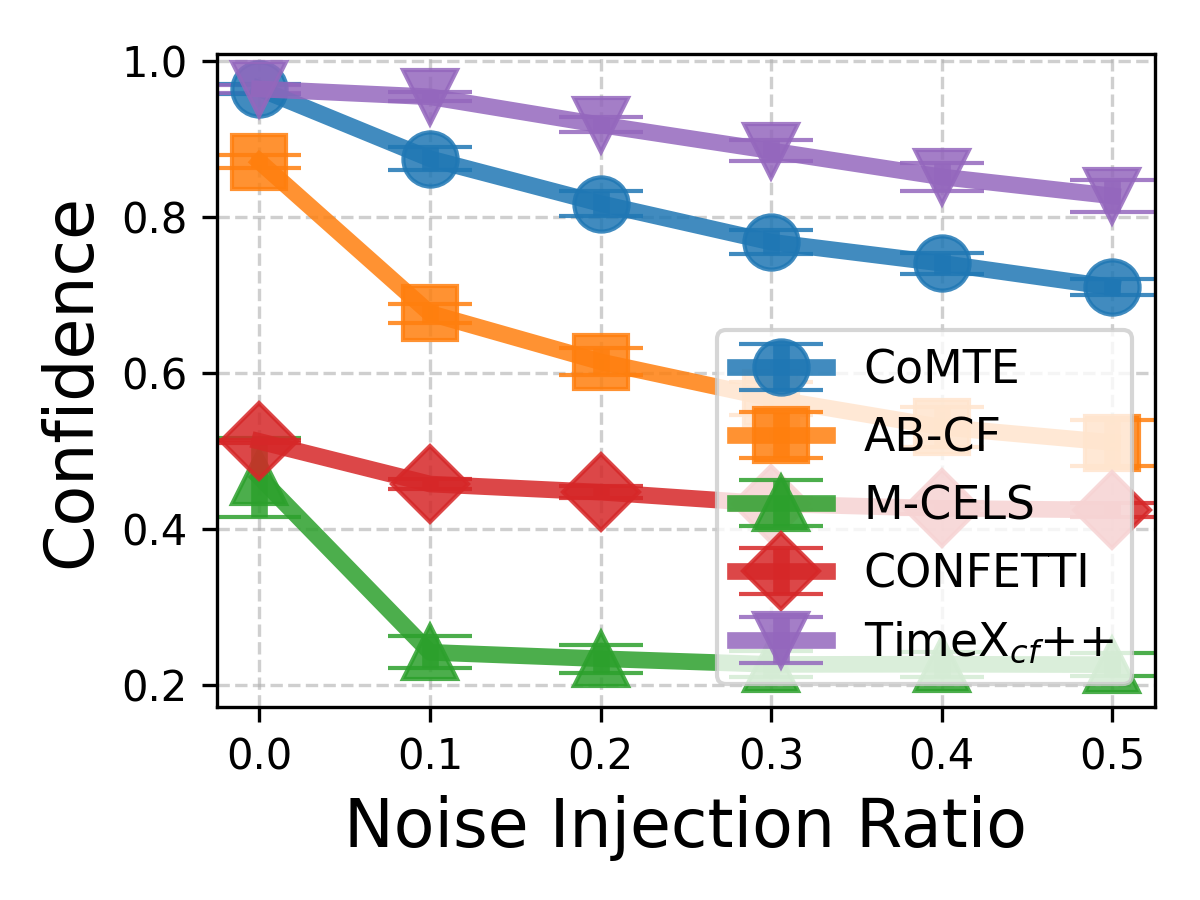}
        \caption{Confidence on Freqshape}
        \label{cf_rob_fig:confidence_freq}
    \end{subfigure}
    
    \begin{subfigure}{0.45\columnwidth}
        \centering
        \includegraphics[width=\linewidth]{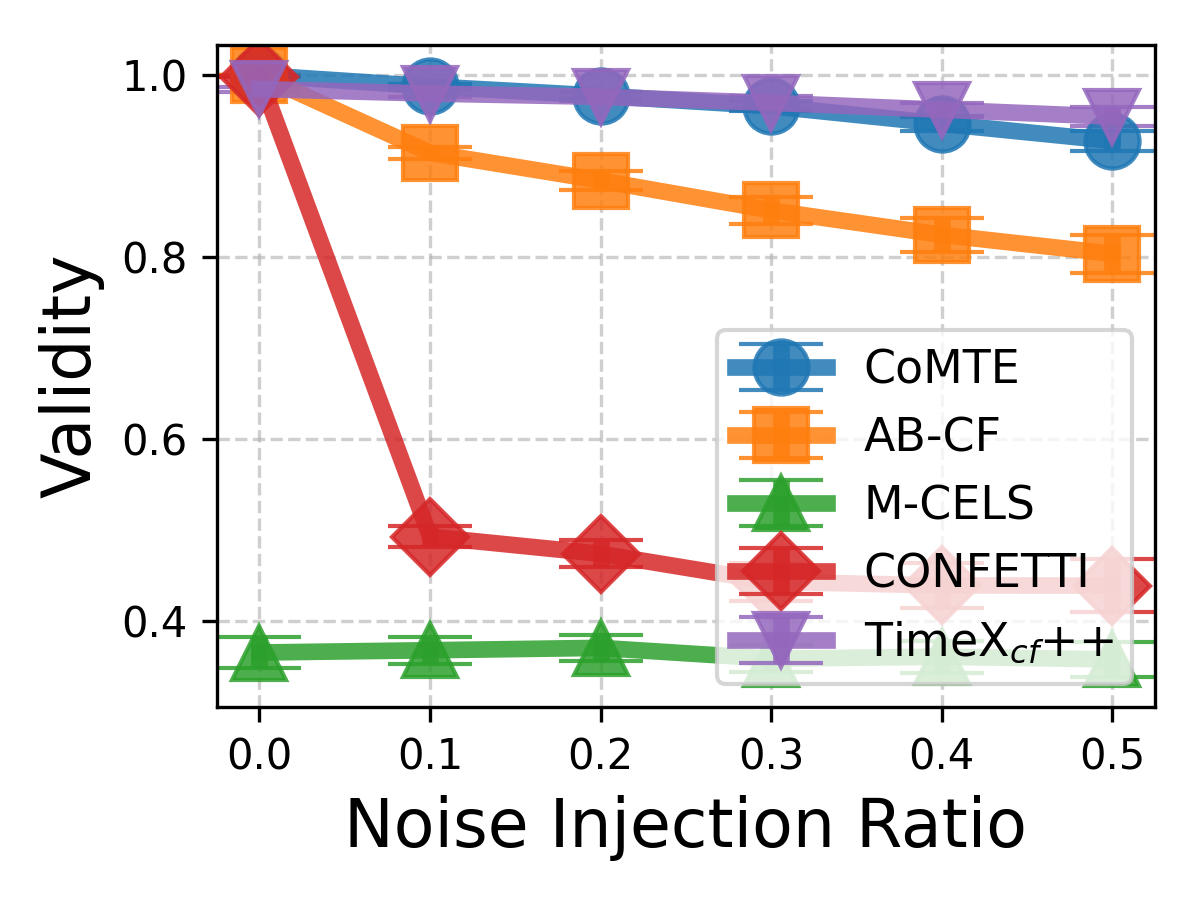}
        \caption{Validity on LowVar}
        \label{cf_rob_fig:val_lowvar}
    \end{subfigure}
    \begin{subfigure}{0.45\columnwidth}
        \centering
        \includegraphics[width=\linewidth]{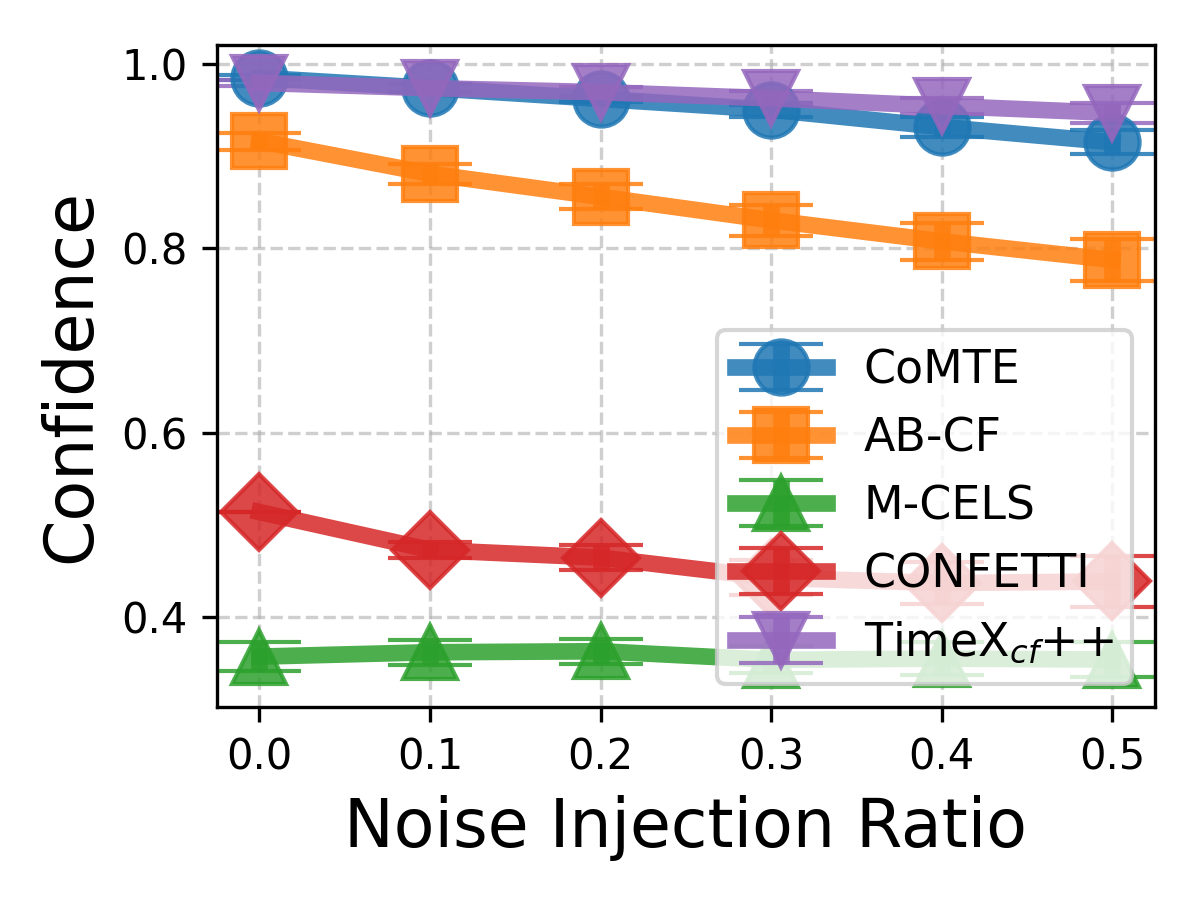}
        \caption{Confidence on LowVar}
        \label{cf_rob_fig:confidence_lowvar}
    \end{subfigure}
	\caption{The counterfactual explanations robustness analysis on Freqshape and LowVar.}
    \vspace{-4mm}
	\label{fig:cf_robustness} 
\end{figure}

Furthermore, because {\ourscf}~strictly confines perturbations to critical structural regions rather than relying on distributed adversarial shortcuts, the resulting counterfactual explanations also exhibit superior robustness to external noise. To evaluate this property, we provide an analysis in Figure~\ref{fig:cf_robustness}. We randomly select timesteps and inject standard Gaussian noise into the generated counterfactual explanations, then assess their effectiveness on the FreqShape and LowVar datasets. The validity and confidence of all counterfactual explanations decrease as the noise injection ratio increases. However, {\ourscf}~demonstrates greater robustness than the baseline methods. Even with a 50\% noise ratio, our counterfactual explanations maintain a validity above 0.8 on both datasets. Although CoMTE achieves comparable robustness, it exhibits much lower sparsity than our method. CONFETTI experiences a sharp drop in validity as noise increases due to its lower confidence. In summary, {\ourscf}~prevents adversarial degeneration by design, demonstrating superior robustness while maintaining better sparsity than the baseline methods.

\begin{table}[t]
    \centering
\setlength{\tabcolsep}{3pt}
    \caption{Ablation of {\oursa}~considering whether there are different losses in our component.}
    \vspace{1mm}\label{tab:ablation_mbc_la}
    \resizebox{0.8\columnwidth}{!}{
    \begin{sc}
    \begin{tabular}{l|ccc}
         \toprule
          & \textbf{AUPRC} & \textbf{AUP} & \textbf{AUR}\\
         \midrule
         
         \multicolumn{4}{c}{ \textbf{FreqShapes}} \\
      \rowcolor{mygreen} {\oursa} & \textbf{0.891}\std{0.002} & \textbf{0.781}\std{0.001} & {0.662}\std{0.002}\\
      w/o  $\mathcal{L}_{\mathrm{LC}}$ & 0.225\std{0.001} & {0.196}\std{0.001} & \textbf{0.768}\std{0.002}\\
       \rowcolor{gray!15} w/o STE  & 0.779\std{0.003} & 0.732\std{0.003} & 0.656\std{0.002} \\
      w/o $\mathcal{L}_{\mathrm{KL}}$ & 0.830\std{0.003} &  0.646\std{0.003} &0.703\std{0.002}\\
       \rowcolor{gray!15} w/o $\mathcal{L}_{dr}$  &  0.194\std{0.002} &  0.139\std{0.002} & 0.415\std{0.002}\\
         \hline
         \hline
         \multicolumn{4}{c}{ \textbf{SeqComb-MV}} \\
       \rowcolor{mygreen} {\oursa} & \textbf{0.759}\std{0.001} & \textbf{0.878}\std{0.001} & {0.391}\std{0.001}\\
            w/o  $\mathcal{L}_{\mathrm{LC}}$ & 0.095\std{0.003} & 0.057\std{0.002} & 0.330\std{0.008}\\
          \rowcolor{gray!15}  w/o STE  & 0.727\std{0.003} & 0.873\std{0.001} & 0.348\std{0.002} \\
           w/o $\mathcal{L}_{\mathrm{KL}}$ & 0.748\std{0.002} &  0.869\std{0.001} & 0.353\std{0.001}\\
           \rowcolor{gray!15}  w/o $\mathcal{L}_{dr}$  & 0.069\std{0.002} & 0.053\std{0.001} & \textbf{0.628}\std{0.008}\\
         \hline
         \hline
         \multicolumn{4}{c}{\textbf{ECG}} \\
        \rowcolor{mygreen}  {\oursa} & \textbf{0.660}\std{0.001} & \textbf{0.726}\std{0.001} & \textbf{0.460}\std{0.001}\\
          w/o STE  & 0.615\std{0.001} & \textbf{0.747}\std{0.001} & 0.402\std{0.001} \\
          \rowcolor{gray!15}  w/o  $\mathcal{L}_{\mathrm{LC}}$ & {0.621}\std{0.002} & {0.642}\std{0.002} & 0.429\std{0.002}\\
           w/o $\mathcal{L}_{\mathrm{KL}}$ & 0.642\std{0.002} & 0.698\std{0.001} & 0.442\std{0.001}\\
          \rowcolor{gray!15}  w/o $\mathcal{L}_{dr}$  & 0.152\std{0.000} &  0.141\std{0.000} & 0.631\std{0.001}\\
         \bottomrule
    \end{tabular}
    \end{sc}
    }
    \vspace{-2mm}
\end{table}

\subsection{Runtime Results}
\label{sec:exp_runtime}
We conduct a runtime analysis evaluating both the training and inference phases across the two explanation paradigms. Regarding the training phase, the runtime for each experiment approximated 3 to 15 minutes per fold, depending on the dataset volume. For attribution explanations, compared to \textsc{TimeX}, which requires training both a white-box model for consistency and the explanation masks, {\oursa}~directly perturbs the black-box, thereby requiring less overall training time. For counterfactual explanations, while {\ourscf}~requires an upfront training stage unlike instance-specific optimization baselines, training time remains highly manageable. It is a constant multiple of the attribution training time, with the factor determined by the size of the classification label space.

During the inference phase, as shown in Table~\ref{times} and Table~\ref{tab:runtime_cf}, \modelname~emerges as the most expedient model across both tasks. For \textbf{attribution}, {\oursa}~is significantly faster than Dynamask and IG, which necessitate recursive operations for individual samples, and slightly faster than \textsc{TimeX}, which requires Landmark calculations. For \textbf{counterfactual} generation, {\ourscf}~demonstrates exceptional efficiency, being over 100 times faster than the fastest baseline (AB-CF). Traditional counterfactual baselines typically rely on costly iterative search procedures to identify suitable perturbations for each new sample. In contrast, because {\ourscf}~is an inductive explainer, it requires only a single forward inference pass and fully benefits from GPU acceleration. Consequently, by requiring only a fixed amount of computation per instance during inference, our framework offers significant practical advantages for large-scale and real-time deployment scenarios.

\subsection{Flexibility Across Different Black-box Classifiers}
\label{sec:exp_classifiers}
To explore the flexibility of \modelname, we study two other time series classifiers and explore their explanatory role. We replace the original transformer-based black-box $f$ with a long-short term memory~(LSTM) and a convolutional neural network (CNN) as the underlying classifiers.

\begin{table*}[t]
    \centering
    \setlength{\tabcolsep}{3pt}
    \caption{{Targeted counterfactual explanation performance on LSTM and CNN architectures.}}
    \resizebox{2.0\columnwidth}{!}{
    \begin{sc}
    \begin{tabular}{c|ccccc|ccccc}
         \toprule
         & \multicolumn{5}{c|}{\textbf{LSTM}} & \multicolumn{5}{c}{\textbf{CNN}} \\
         \textbf{Method} & \textbf{Validity} & \textbf{Confidence} & \textbf{Sparsity}  & \textbf{Prox.-L1} & \textbf{Prox.-L2} & \textbf{Validity} & \textbf{Confidence} & \textbf{Sparsity}  & \textbf{Prox.-L1} & \textbf{Prox.-L2} \\
         \midrule
          \multicolumn{11}{c}{\textbf{Freqshapes}}\\
          \rowcolor{gray!15} CoMTE &  1.000 \std{ 0.000	}&	0.935 \std{ 0.014	}&	1.000 \std{ 0.000	}&	0.562 \std{ 0.003	}&	0.821 \std{ 0.006	}& 1.000 \std{ 0.000	}&	0.922 \std{ 0.010	}&	1.000 \std{ 0.000	}&	0.578 \std{ 0.002	}&	0.847 \std{ 0.003	} \\
         AB-CF & 1.000 \std{ 0.000	}&	0.823 \std{ 0.017	}&	0.664 \std{ 0.021	}&	0.424 \std{ 0.007	}&	0.723 \std{ 0.005	}&1.000 \std{ 0.000	}&	0.790 \std{ 0.010	}&	0.715 \std{ 0.015	}&	0.451 \std{ 0.006	}&	0.760 \std{ 0.005	} \\
          \rowcolor{gray!15} M-CELS & 0.294 \std{ 0.029	}&	0.267 \std{ 0.026	}&	0.057 \std{ 0.006	}&	0.070 \std{ 0.008	}&	0.192 \std{ 0.018	}&0.405 \std{ 0.025	}&	0.363 \std{ 0.022	}&	0.055 \std{ 0.003	}&	0.072 \std{ 0.005	}&	0.207 \std{ 0.013	}\\
         CONFETTI & 0.981 \std{ 0.008	}&	0.506 \std{ 0.004	}&	0.361 \std{ 0.005	}&	0.260 \std{ 0.004	}&	0.568 \std{ 0.007	}&0.984 \std{ 0.010	}&	0.507 \std{ 0.005	}&	0.360 \std{ 0.005	}&	0.275 \std{ 0.004	}&	0.612 \std{ 0.006	}\\
         \cmidrule(r){1-11}
          \rowcolor{mygreen} {\ourscf} & 0.944 \std{ 0.036	}&	0.926 \std{ 0.037	}&	0.271 \std{ 0.011	}&	0.421 \std{ 0.015	}&	0.821 \std{ 0.022	}&0.976 \std{ 0.007	}&	0.967 \std{ 0.007	}&	0.283 \std{ 0.014	}&	0.564 \std{ 0.031	}&	1.120 \std{ 0.089	}\\
         \midrule
         \midrule
          \multicolumn{11}{c}{\textbf{ SeqComb-MV}}\\
          \rowcolor{gray!15} CoMTE &  0.992 \std{ 0.003	}&	0.932 \std{ 0.018	}&	0.425 \std{ 0.020	}&	0.078 \std{ 0.003	}&	0.239 \std{ 0.007	}&0.999 \std{ 0.001	}&	0.969 \std{ 0.008	}&	0.454 \std{ 0.017	}&	0.085 \std{ 0.003	}&	0.255 \std{ 0.007	}\\
         AB-CF & 1.000 \std{ 0.000	}&	0.862 \std{ 0.025	}&	0.603 \std{ 0.033	}&	0.094 \std{ 0.004	}&	0.242 \std{ 0.007	}& 1.000 \std{ 0.000	}&	0.900 \std{ 0.015	}&	0.578 \std{ 0.037	}&	0.094 \std{ 0.005	}&	0.249 \std{ 0.008	} \\
          \rowcolor{gray!15} M-CELS & 0.284 \std{ 0.039	}&	0.277 \std{ 0.036	}&	0.011 \std{ 0.001	}&	0.007 \std{ 0.001	}&	0.064 \std{ 0.006	}& 0.549 \std{ 0.059	}&	0.516 \std{ 0.056	}&	0.012 \std{ 0.001	}&	0.011 \std{ 0.001	}&	0.089 \std{ 0.008	}\\
         CONFETTI & 0.911 \std{ 0.042	}&	0.504 \std{ 0.021	}&	0.236 \std{ 0.017	}&	0.036 \std{ 0.003	}&	0.140 \std{ 0.011	}& 0.998 \std{ 0.001	}&	0.512 \std{ 0.000	}&	0.304 \std{ 0.011	}&	0.051 \std{ 0.003	}&	0.189 \std{ 0.010	}\\
          
         \cmidrule(r){1-11}
          \rowcolor{mygreen} {\ourscf} & 0.941 \std{ 0.037	}&	0.942 \std{ 0.034	}&	0.129 \std{ 0.014	}&	0.121 \std{ 0.013	}&	0.364 \std{ 0.023	}&0.940 \std{ 0.035	}&	0.941 \std{ 0.032	}&	0.127 \std{ 0.013	}&	0.124 \std{ 0.015	}&	0.374 \std{ 0.028	}\\
         \midrule
         \midrule
          \multicolumn{11}{c}{\textbf{ECG}}\\
          \rowcolor{gray!15} CoMTE &  1.000 \std{ 0.000	}&	0.852 \std{ 0.030	}&	1.000 \std{ 0.000	}&	0.279 \std{ 0.010	}&	0.384 \std{ 0.016	}& 1.000 \std{ 0.000	}&	0.873 \std{ 0.016	}&	1.000 \std{ 0.000	}&	0.292 \std{ 0.002	}&	0.403 \std{ 0.003	}\\
         AB-CF & 1.000 \std{ 0.000	}&	0.779 \std{ 0.036	}&	0.570 \std{ 0.057	}&	0.174 \std{ 0.016	}&	0.303 \std{ 0.019	}& 0.999 \std{ 0.000	}&	0.750 \std{ 0.015	}&	0.464 \std{ 0.047	}&	0.165 \std{ 0.006	}&	0.317 \std{ 0.010	}\\
          \rowcolor{gray!15} M-CELS & 0.368 \std{ 0.057	}&	0.358 \std{ 0.056	}&	0.015 \std{ 0.001	}&	0.007 \std{ 0.001	}&	0.042 \std{ 0.004	}& 0.486 \std{ 0.063	}&	0.462 \std{ 0.062	}&	0.023 \std{ 0.002	}&	0.015 \std{ 0.002	}&	0.077 \std{ 0.007	}\\
         CONFETTI & 0.858 \std{ 0.030	}&	0.485 \std{ 0.020	}&	0.250 \std{ 0.010	}&	0.072 \std{ 0.005	}&	0.180 \std{ 0.013	}& 0.977 \std{ 0.007	}&	0.503 \std{ 0.004	}&	0.255 \std{ 0.010	}&	0.076 \std{ 0.005	}&	0.208 \std{ 0.012	}\\
         \cmidrule(r){1-11}
          \rowcolor{mygreen} {\ourscf} & 0.772 \std{ 0.062	}&	0.738 \std{ 0.057	}&	0.138 \std{ 0.020	}&	0.251 \std{ 0.037	}&	0.709 \std{ 0.096	}& 0.853 \std{ 0.047	}&	0.849 \std{ 0.047	}&	0.049 \std{ 0.016	}&	0.104 \std{ 0.008	}&	0.542 \std{ 0.033	}\\
         \bottomrule
    \end{tabular}
    \end{sc}
    }
    \label{tab:cf_models}
\end{table*}
\begin{table}[!h]
    \centering
\setlength{\tabcolsep}{2pt}
    \caption{{Ablation study Counterfactual explanation performance on LSTM and CNN architectures.}}
    \resizebox{1.0\columnwidth}{!}{
    \begin{sc}
    \begin{tabular}{l|ccccc}
         \toprule
         & \textbf{Validity} & \textbf{Confidence} & \textbf{Sparsity}  & \textbf{Prox.-L1} & \textbf{Prox.-L2} \\
         \midrule
          \multicolumn{6}{c}{\textbf{Freqshapes}} \\
         \rowcolor{mygreen} {\ourscf} &  {0.974} \std{ {0.004}}&	{0.964}\std{{0.005}}&	0.214 \std{ 0.005	}&	0.339 \std{ 0.021	}&	0.757 \std{ 0.030	} \\
         w/o STE & 0.769 \std{ 0.055	}&	0.763 \std{ 0.055	}&	0.174 \std{ 0.014	}&	0.281 \std{ 0.031	}&	0.616 \std{ 0.059	} \\
         \rowcolor{gray!15} w/o  $\mathcal{L}_{\mathrm{LC}}$  & 0.000 \std{ 0.000	}&	0.007 \std{ 0.001	}&	0.091 \std{ 0.001	}&	0.001 \std{ 0.000	}&	0.002 \std{ 0.000	} \\
         w/o $\mathcal{L}_{\mathrm{KL}}$ & 0.936 \std{ 0.035	}&	0.928 \std{ 0.034	}&	0.220 \std{ 0.012	}&	0.342 \std{ 0.030	}&	0.730 \std{ 0.048	} \\
         \rowcolor{gray!15} w/o $\mathcal{L}_{dr}$ & 0.955 \std{ 0.032	}&	0.946 \std{ 0.032	}&	0.216 \std{ 0.014	}&	0.512 \std{ 0.070	}&	1.107 \std{ 0.128	} \\
         \midrule
         \midrule
          \multicolumn{6}{c}{\textbf{SeqComb-MV}} \\
        \rowcolor{mygreen}  {\ourscf} & {0.983} \std{{0.007}}& {0.979} \std{{0.008}}&	0.305 \std{ 0.041	}&	0.152 \std{ 0.016	}&	0.353 \std{ 0.033	} \\
         w/o STE & 0.804 \std{ 0.049	}&	0.799 \std{ 0.049	}&	0.114 \std{ 0.007	}&	0.114 \std{ 0.016	}&	0.368 \std{ 0.030	} \\
          \rowcolor{gray!15} w/o  $\mathcal{L}_{\mathrm{LC}}$  & 0.000 \std{ 0.000	}&	0.003 \std{ 0.001	}&	0.079 \std{ 0.005	}&	0.000 \std{ 0.000 	}&	0.001 \std{ 0.000	} \\
         w/o $\mathcal{L}_{\mathrm{KL}}$ & 0.940 \std{ 0.036	}&	0.941 \std{ 0.033	}&	0.150 \std{ 0.015	}&	0.146 \std{ 0.016	}&	0.395 \std{ 0.025	} \\
          \rowcolor{gray!15} w/o $\mathcal{L}_{dr}$ & 0.907 \std{ 0.039	}&	0.907 \std{ 0.038	}&	0.129 \std{ 0.013	}&	0.286 \std{ 0.062	}&	0.771 \std{ 0.122	} \\
         \midrule
         \midrule
          \multicolumn{6}{c}{\textbf{ECG}} \\
        \rowcolor{mygreen}  {\ourscf} &  0.848 \std{ 0.104	}&	0.847 \std{ 0.101	}&	0.084 \std{ 0.026	}&	0.251 \std{ 0.097	}&	0.767 \std{ 0.185	} \\
        w/o STE & 0.637 \std{ 0.128	}&	0.638 \std{ 0.128	}&	0.049 \std{ 0.015	}&	0.111 \std{ 0.031	}&	0.499 \std{ 0.080	}\\
         \rowcolor{gray!15} w/o  $\mathcal{L}_{\mathrm{LC}}$  & 0.000 \std{ 0.000	}&	0.015 \std{ 0.005	}&	0.096 \std{ 0.003	}&	0.000 \std{ 0.000	}&	0.001 \std{ 0.000	} \\
         w/o $\mathcal{L}_{\mathrm{KL}}$ & 0.824 \std{ 0.124	}&	0.831 \std{ 0.116	}&	0.104 \std{ 0.037	}&	0.311 \std{ 0.139	}&	0.850 \std{ 0.214	}\\
          \rowcolor{gray!15} w/o $\mathcal{L}_{dr}$ & 0.847 \std{ 0.110	}&	0.853 \std{ 0.102	}&	0.106 \std{ 0.013	}&	0.539 \std{ 0.084	}&	1.675 \std{ 0.221	} \\
         \bottomrule
    \end{tabular}
    \end{sc}
    }
    \vspace{-4mm}
    \label{tab:cf_ablation}
\end{table}

For \textbf{attribution} explanations, {\oursa}~retains the best AUPRC prediction on these datasets and is also slightly ahead for AUP and AUR overall, as shown in Table~\ref{table:lstm_synth}, while \textsc{TimeX} fails to predict on SeqComb-MV due to non-convergence. Under the CNN predictor, our method achieves the highest AUPRC and AUP for both SeqComb-MV and ECG datasets. 

For \textbf{counterfactual} explanations, Table~\ref{tab:cf_models} presents the comparison results on three benchmarks. For the baselines, similar behaviors are observed across different architectures, except for M-CELS, whose performance exhibits large variance, performing better with CNN and worse with LSTM. We also observe results consistent with our main findings, where {\ourscf}~achieves a balance among validity, confidence, sparsity, and proximity across both CNN and LSTM backbones, indicating strong generalizability across architectures.

\subsection{Ablation Study and Hyperparameter Analysis}
\label{sec:exp_ablation}

To disentangle the contributions of the proposed components, we conduct extensive ablation studies and hyperparameter analyses across both attribution and counterfactual tasks, evaluated on FreqShape, SeqComb-MV, and ECG datasets.

\paragraph{Effect of STE} 
For \textbf{attribution} explanations as shown in Table~\ref{tab:ablation_mbc_la}, {\oursa}~outperforms leading baselines by $39.78\%$ in AUPRC, $22.04\%$ in AUP, and $3.10\%$ in AUR on ECG arrhythmia detection, indicating that the use of the STE plays a critical role in optimizing the discrete explanation masks. On average, employing STE yields an average $8.87\%$ increase in AUPRC across all datasets compared to a continuous masking strategy, alongside consistent enhancements in AUR, providing tangible evidence for utilizing hard masks.  
In the \textbf{counterfactual} domain, removing the STE results in a significant drop in validity and confidence. Although sparsity superficially improves, this indicates that the generation process has collapsed into a local optimum where perturbations are overly constrained and fail to effectively flip the model prediction, highlighting the necessity of STE in balancing the multi-objective trade-off.

\begin{figure}[t]
	\centering
    \begin{subfigure}{0.31\linewidth}
        \centering
        \includegraphics[width=\linewidth]{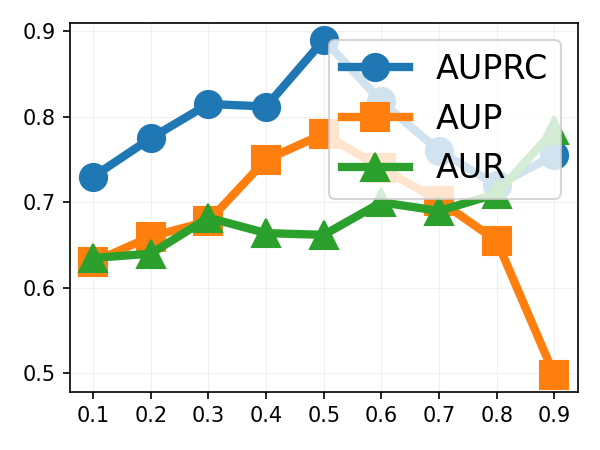}
        \caption{Freqshape}
    \end{subfigure}
    \begin{subfigure}{0.31\linewidth}
        \centering
        \includegraphics[width=\linewidth]{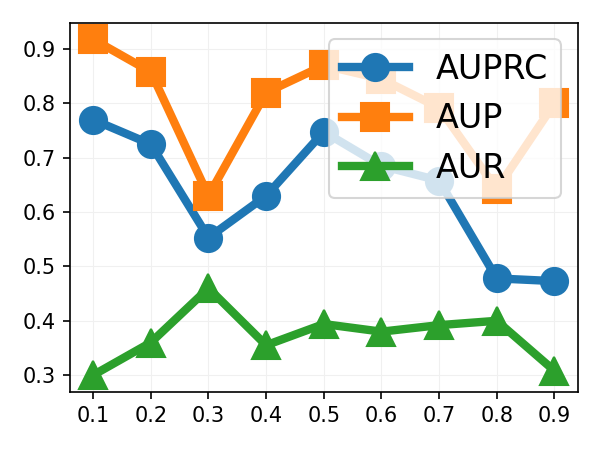}
        \caption{SeqComb-MV}
    \end{subfigure}
    \begin{subfigure}{0.31\linewidth}
        \centering
        \includegraphics[width=\linewidth]{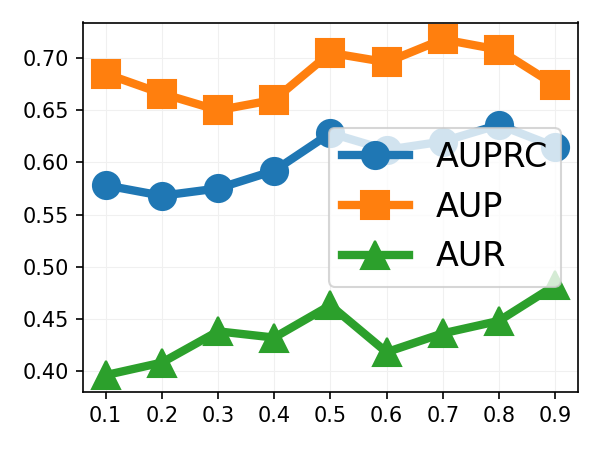}
        \caption{ECG}
        \label{stab3}
    \end{subfigure}
	\caption{\textbf{Attribution} explanation performance varying the sparsity parameter $r$.}
    \label{rvalues}
    \vspace{-4mm}
\end{figure}

\begin{figure*}[t]	
	\centering
    \begin{subfigure}{0.195\linewidth}
        \centering
        \includegraphics[width=\linewidth]{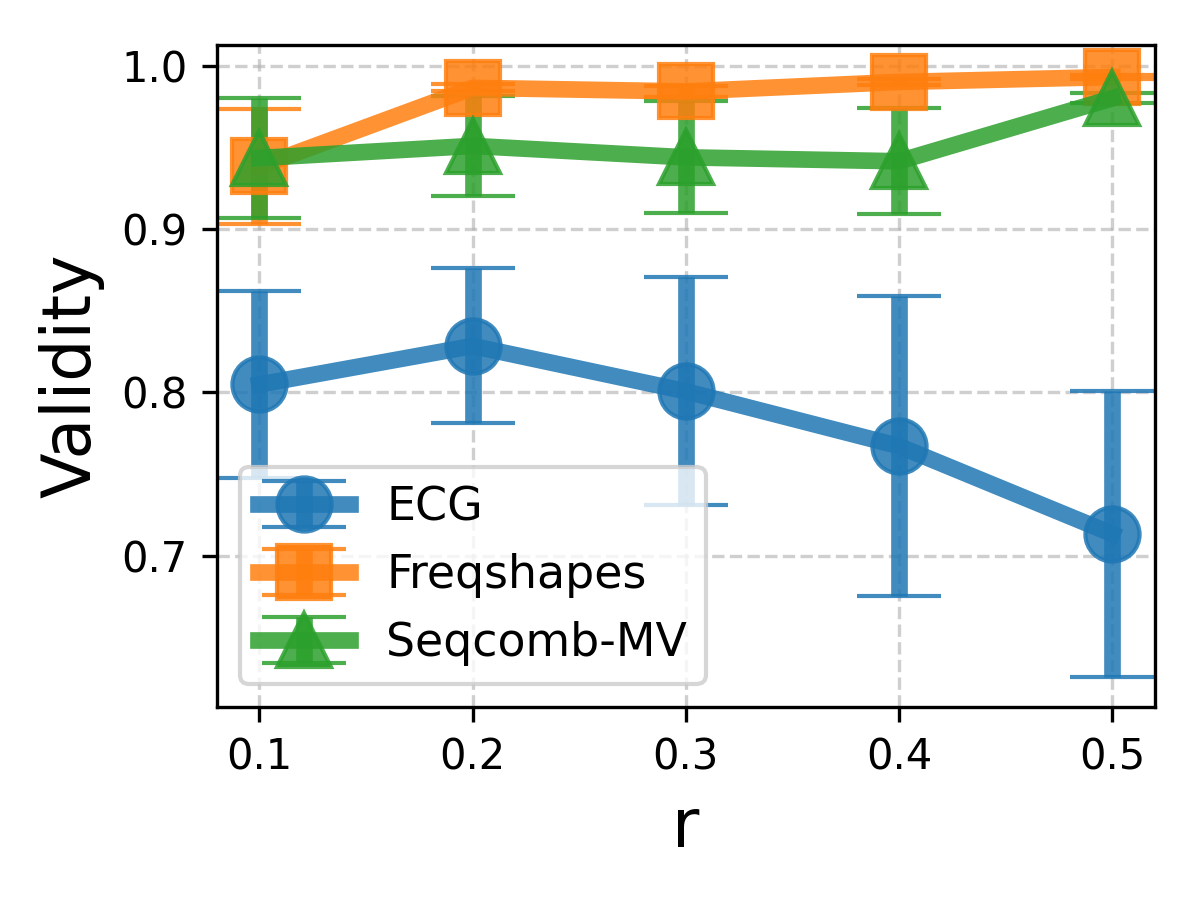}
        \caption{Validity}
    \end{subfigure}
    \begin{subfigure}{0.195\linewidth}
        \centering
        \includegraphics[width=\linewidth]{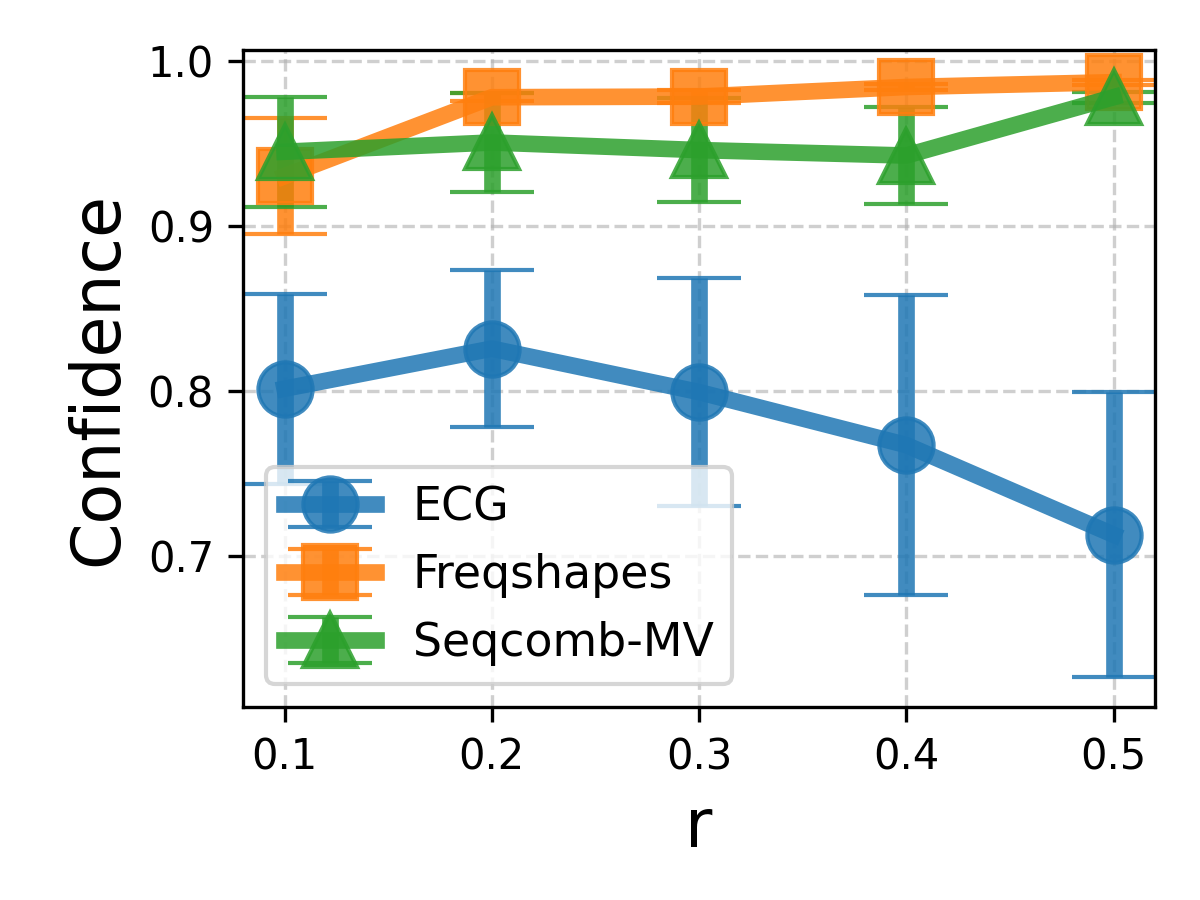}
        \caption{Confidence}
    \end{subfigure}
    \begin{subfigure}{0.195\linewidth}
        \centering
        \includegraphics[width=\linewidth]{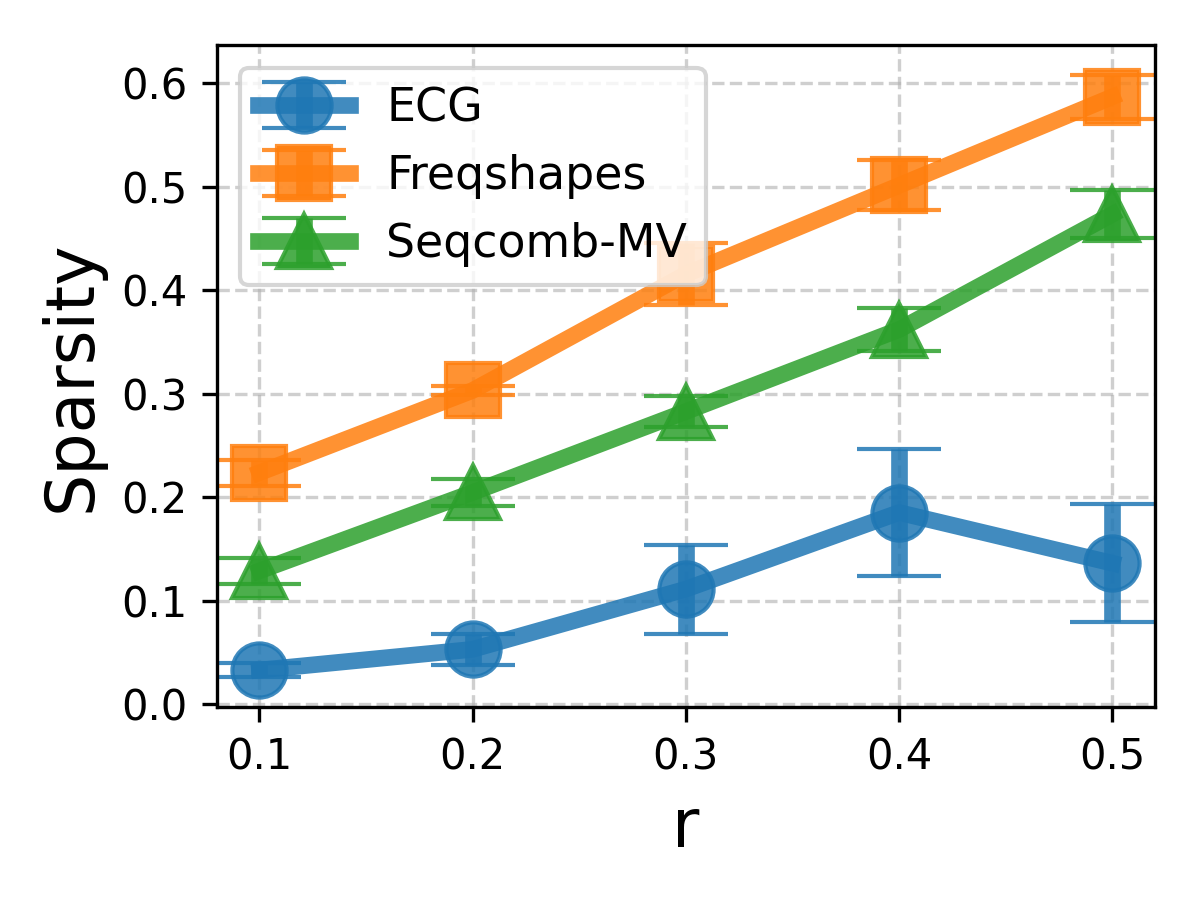}
        \caption{Sparsity}
    \end{subfigure}
    \begin{subfigure}{0.195\linewidth}
        \centering
        \includegraphics[width=\linewidth]{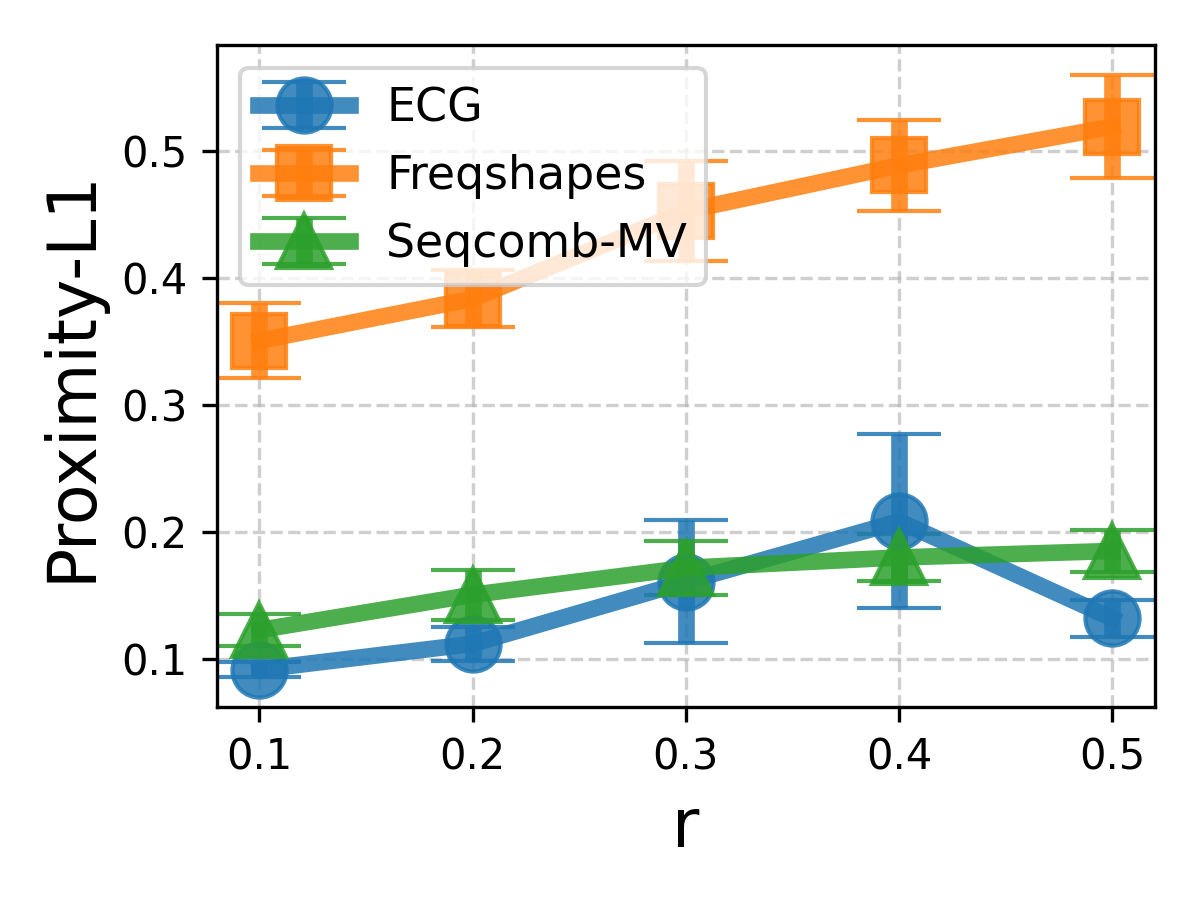}
        \caption{Proximity-L1}
    \end{subfigure}
    \begin{subfigure}{0.195\linewidth}
        \centering
        \includegraphics[width=\linewidth]{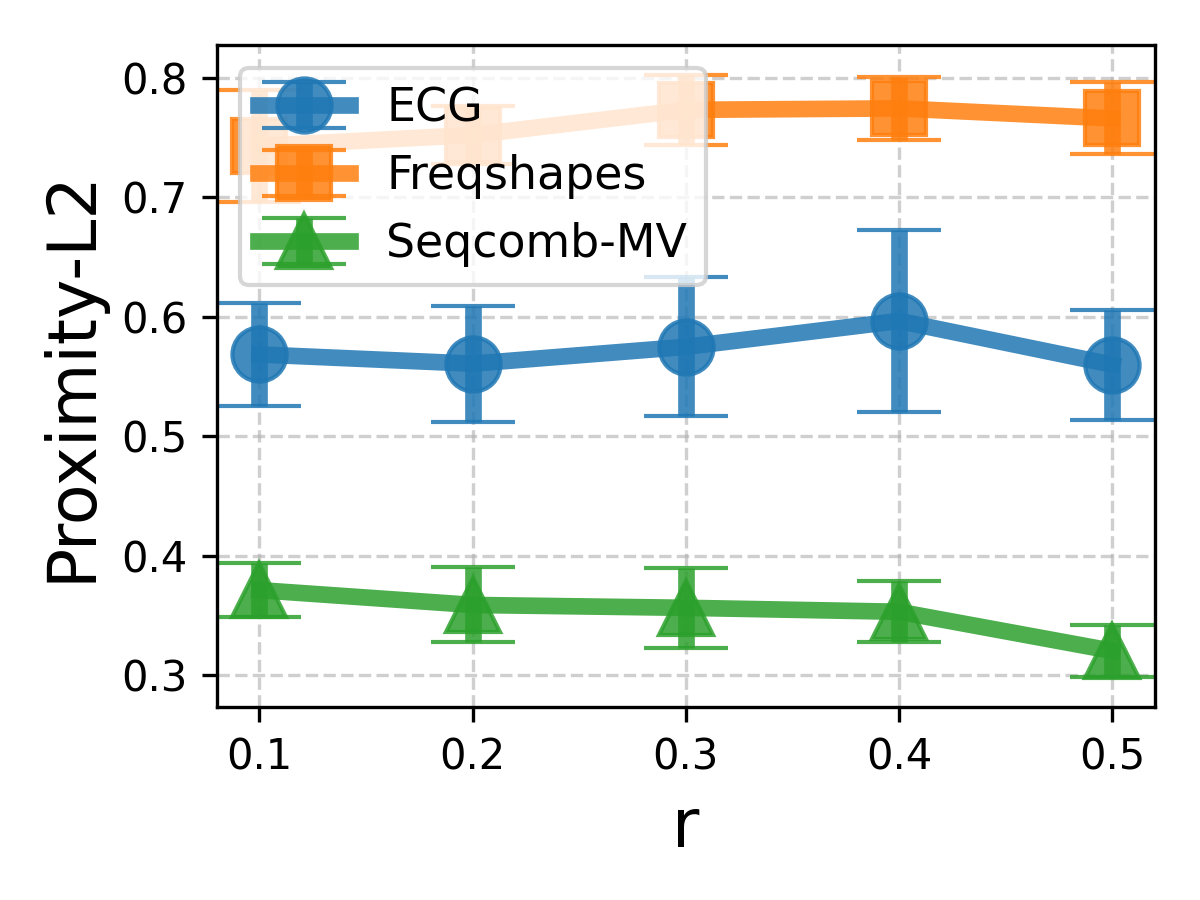}
        \caption{Proximity-L2}
    \end{subfigure}
	\caption{The hyperparameter $r$ analysis for \textbf{counterfactual} explanations on Freqshape (univariate), SeqComb-MV (multivariate), and ECG (real-world) datasets.}
	\label{fig:cf_r} 
    \vspace{-4mm}
\end{figure*}

\paragraph{Effect of Different Losses} 
We examine the effectiveness of the individual loss components constraining our framework. For \textbf{attribution} explanations, the absence of the Kullback-Leibler maintenance loss ($\mathcal{L}_{\mathrm{KL}}$) widens the distributional gap between original and perturbed instances, dropping performance. 
The removal of the reference distance loss ($\mathcal{L}_{dr}$) naturally causes the explainer to fail, as it removes the primary target for bottleneck extraction. Furthermore, without consistency labeling ($\mathcal{L}_{\mathrm{LC}}$), performance on synthetic datasets drops significantly.

\begin{figure}[t]
    \centering
    \includegraphics[width=0.80\columnwidth]{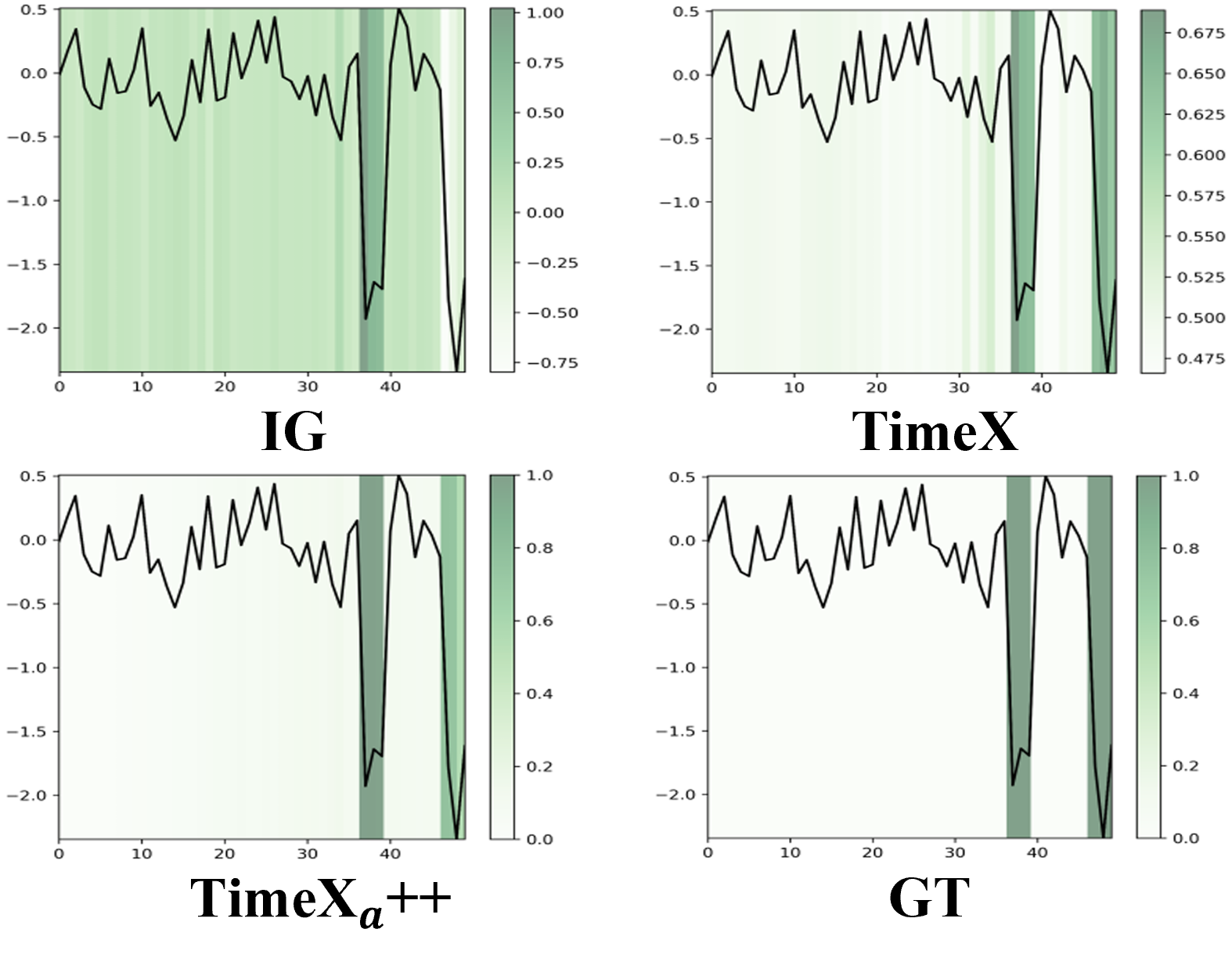}
    \caption{Visualization of two attribution explainers and {\oursa} on the FreqShapes dataset. }
    \label{fig:vis_methods}
    \vspace{-4mm}
\end{figure}

In the \textbf{counterfactual} domain (Table~\ref{tab:cf_ablation}), removing the label consistency loss ($\mathcal{L}_{\text{LC}}$) dramatically degrades validity, as the generator loses its primary signal for targeted label flipping. 
Removing the reference distance loss ($\mathcal{L}_{dr}$), which acts as a structural regularization term, leads to a substantial decline in proximity. 
The maintenance loss ($\mathcal{L}_{\mathrm{KL}}$) has a relatively smaller but consistent impact on preserving distributional consistency. 
Interestingly, the effect of the connectivity loss ($\mathcal{L}_{\text{con}}$) varies depending on the underlying data characteristics, which improves performance on FreqShapes by promoting continuous modifications but degrades performance on ECG, as critical ECG patterns are inherently distributed across multiple discrete segments rather than concentrated in a single contiguous window. 
Finally, as previously analyzed, excising the structural anchor ($\mathcal{L}_{bound}$) results in a dramatic deterioration of sparsity and proximity metrics; without this causal anchor, the generator scatters perturbations across the entire sequence.

\paragraph{Choosing the Hyperparameter $r$} The parameter $r$ governs the sparsity of the masks learned during the process. For \textbf{attribution}, in Figure~\ref{rvalues}, lower values of the parameter are associated with a decrease in the performance (AUR). The performance stabilizes when the value is between $0.4$ and $0.7$, suggesting robustness in this range. In Figure~\ref{fig:cf_r}, for \textbf{counterfactuals}, $r$ balances sparsity, validity, and proximity by providing prior knowledge about the expected size of counterfactual explanations. As $r$ increases, validity and confidence improve, whereas sparsity and proximity deteriorate, reflecting the inherent multi-objective trade-offs.

\begin{figure}[t]	
	\centering
    \begin{subfigure}{1.0\linewidth}
        \centering
        \includegraphics[width=\linewidth]{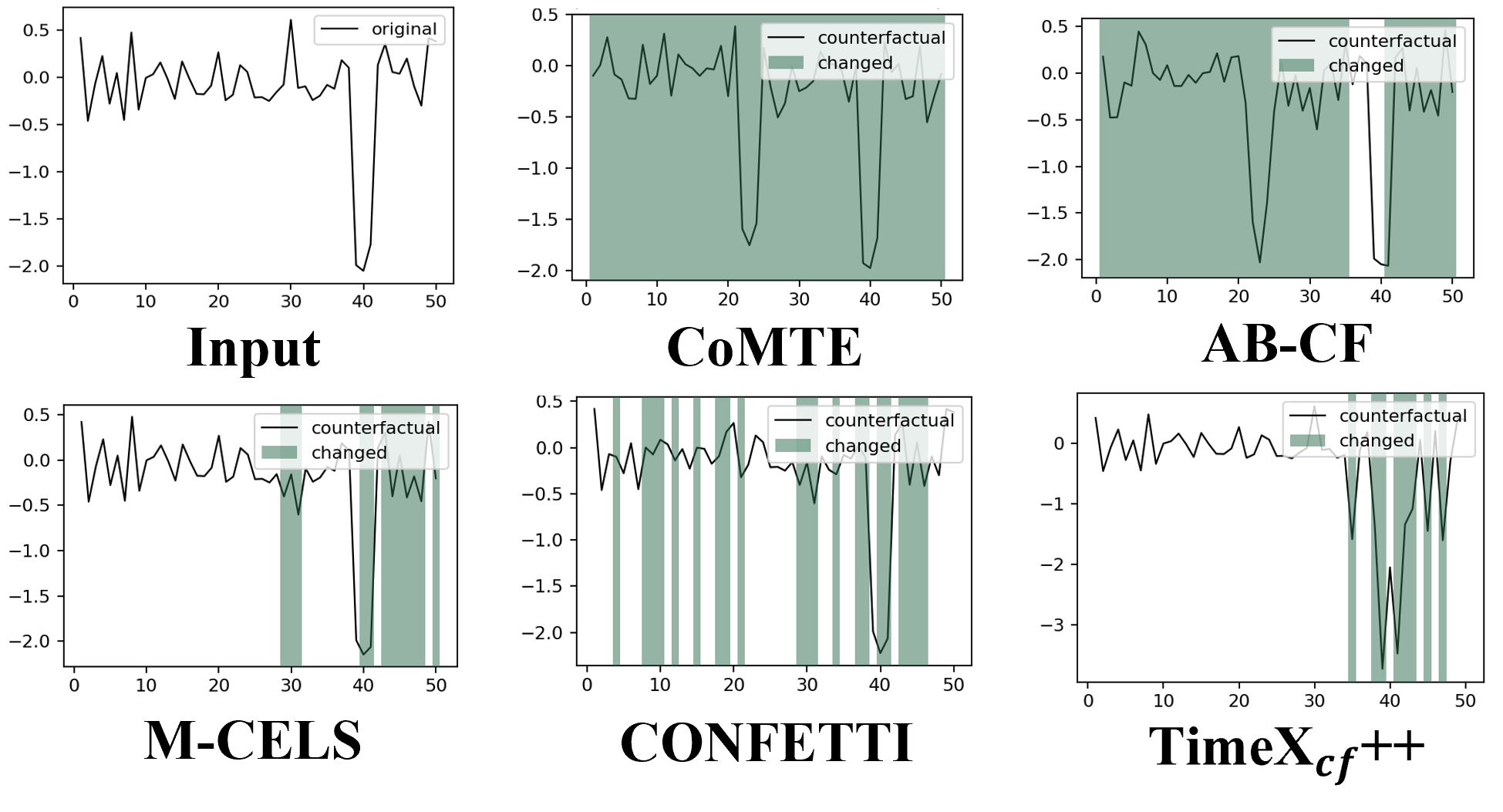}
    \end{subfigure}
	\caption{The cases visualization of the FreqShapes dataset for counterfactual explanations. We demonstrate pairs of counterfactual cases with prediction 2 and target label 0. } 
	\label{fig:cf_case} 
    \vspace{-4mm}
\end{figure}

\subsection{Case Study}
\label{sec:exp_casestudy} 
Saliency maps are a potent tool for visualizing the significance of features. We demonstrate the attribution saliency maps of the benchmarks and {\oursa}~on the FreqShapes dataset in Figure~\ref{fig:vis_methods}. IG identifies unnecessarily large areas as important, which can be untenable for noisy datasets. \textsc{TimeX} struggles to describe important sub-instances with certainty due to data distribution bias from its constructed classifiers. In stark contrast, {\oursa}~focuses accurately on the peaks for matching ground-truth explanations.

For counterfactual explanations in Figure~\ref{fig:cf_case}, we compare {\ourscf}~with baseline methods, with the modified parts highlighted. {\ourscf}~requires fewer modifications, and these are highly related to the underlying patterns. In contrast, CoMTE and AB-CF modify a large portion of the original input, resulting in trivial solutions (e.g., generating highly similar instances). M-CELS fails to generate valid counterfactual explanations in most cases. CONFETTI produces sparse modifications but its results lack robustness to noise.

\section{Conclusion}
In this work, we theoretically investigate an information-theoretic guided objective for time series explanations that ensures compactness and informativeness.
We propose a novel approach, \modelname, based on the IB principle, which allows a traceability computation to produce faithful attribution explanations and robust counterfactual explanations.
Comprehensive studies on synthetic and real-world datasets have confirmed that
\modelname~surpasses existing approaches in performance and efficiency. 
Its effectiveness shows that \modelname's capability to reflect complex behaviors of pre-trained time series classifiers accurately.
However, it may involve some hyperparameters in the learning objective to control the quantifiers of the explanation, especially when dealing with different datasets, which might be the key limitation.

\bibliographystyle{plain}
\bibliography{ref}

@inproceedings{tishby2015deep,
  title={Deep learning and the information bottleneck principle},
  author={Tishby, Naftali and Zaslavsky, Noga},
  booktitle={2015 IEEE Information Theory Workshop},
  pages={1--5},
  year={2015},
  organization={IEEE}
}

@inproceedings{miao2022interpretable,
  title={Interpretable and generalizable graph learning via stochastic attention mechanism},
  author={Miao, Siqi and Liu, Mia and Li, Pan},
  booktitle={ICML},
  pages={15524--15543},
  year={2022},
  organization={PMLR}
}

@article{luo2020parameterized,
  title={Parameterized explainer for graph neural network},
  author={Luo, Dongsheng and Cheng, Wei and Xu, Dongkuan and Yu, Wenchao and Zong, Bo and Chen, Haifeng and Zhang, Xiang},
  journal={NeurIPS},
  volume={33},
  pages={19620--19631},
  year={2020}
}

@inproceedings{delaney2021instance,
  title={Instance-based counterfactual explanations for time series classification},
  author={Delaney, Eoin and Greene, Derek and Keane, Mark T},
  booktitle={International Conference on Case-Based Reasoning},
  pages={32--47},
  year={2021},
  organization={Springer}
}

@inproceedings{queen2023encoding,
  title={Encoding Time-Series Explanations through Self-Supervised Model Behavior Consistency},
  author={Queen, Owen and Hartvigsen, Thomas and Koker, Teddy and He, Huan and Tsiligkaridis, Theodoros and Zitnik, Marinka},
  booktitle={NeurIPS},
  year={2023}
}

@article{geirhos2020shortcut, 
 title={Shortcut learning in deep neural networks},
 volume={2}, 
 number={11}, 
 journal={Nature Machine Intelligence}, 
 author={Geirhos, Robert and Jacobsen, Jörn-Henrik and Michaelis, Claudio and Zemel, Richard and Brendel, Wieland and Bethge, Matthias and Wichmann, Felix A.}, 
 year={2020}, 
 pages={665–673}
 }

@inproceedings{faber2021comparing,
  title={When comparing to ground truth is wrong: On evaluating gnn explanation methods},
  author={Faber, Lukas and K. Moghaddam, Amin and Wattenhofer, Roger},
  booktitle={SIGKDD},
  pages={332--341},
  year={2021}
}

@article{zhang2021survey,
  title={A survey on neural network interpretability},
  author={Zhang, Yu and Ti{\v{n}}o, Peter and Leonardis, Ale{\v{s}} and Tang, Ke},
  journal={IEEE Transactions on Emerging Topics in Computational Intelligence},
  volume={5},
  number={5},
  pages={726--742},
  year={2021},
  publisher={IEEE}
}

@inproceedings{reiss2012introducing,
  title={Introducing a new benchmarked dataset for activity monitoring},
  author={Reiss, Attila and Stricker, Didier},
  booktitle={ISWC},
  pages={108--109},
  year={2012}
}

@article{Moody2001TheIO,
  title={The impact of the {MIT-BIH} Arrhythmia Database},
  author={George B. Moody and Roger G. Mark},
  journal={IEEE Engineering in Medicine and Biology Magazine},
  year={2001},
  volume={20},
  pages={45-50}
}

@article{andrzejak2001indications,
  title={Indications of nonlinear deterministic and finite-dimensional structures in time series of brain electrical activity: Dependence on recording region and brain state},
  author={Andrzejak, Ralph G and Lehnertz, Klaus and Mormann, Florian and Rieke, Christoph and David, Peter and Elger, Christian E},
  journal={Physical Review E},
  pages={061907},
  year={2001}
}

@inproceedings{sundararajan2017axiomatic,
  title={Axiomatic attribution for deep networks},
  author={Sundararajan, Mukund and Taly, Ankur and Yan, Qiqi},
  booktitle={ICML},
  pages={3319--3328},
  year={2017}
}

@inproceedings{crabbe2021explaining,
  title={Explaining time series predictions with dynamic masks},
  author={Crabb{\'e}, Jonathan and Van Der Schaar, Mihaela},
  booktitle={ICML},
  pages={2166--2177},
  year={2021}
}

@inproceedings{leung2023temporal,
title={Temporal Dependencies in Feature Importance for Time Series Prediction},
author={Kin Kwan Leung and Clayton Rooke and Jonathan Smith and Saba Zuberi and Maksims Volkovs},
booktitle={ICLR},
  pages={1--18},
year={2023}
}

@inproceedings{
chuang2023cortx,
title={Co{RTX}: Contrastive Framework for Real-time Explanation},
author={Yu-Neng Chuang and Guanchu Wang and Fan Yang and Quan Zhou and Pushkar Tripathi and Xuanting Cai and Xia Hu},
booktitle={ICLR},
  pages={1--23},
year={2023}
}

@inproceedings{ismail2021improving,
  title={Improving deep learning interpretability by saliency guided training},
  author={Ismail, Aya Abdelsalam and Corrada Bravo, Hector and Feizi, Soheil},
  booktitle={NeurIPS},
  pages={26726--26739},
  year={2021}
}

@inproceedings{vaswani2017attention,
  title={Attention is all you need},
  author={Vaswani, Ashish and Shazeer, Noam and Parmar, Niki and Uszkoreit, Jakob and Jones, Llion and Gomez, Aidan N and Kaiser, {\L}ukasz and Polosukhin, Illia},
  booktitle={NeurIPS},
    pages={5998--6008},
  year={2017}
}

@inproceedings{zhao2022ood,
  title={{OOD-CV}: A benchmark for robustness to out-of-distribution shifts of individual nuisances in natural images},
  author={Zhao, Bingchen and Yu, Shaozuo and Ma, Wufei and Yu, Mingxin and Mei, Shenxiao and Wang, Angtian and He, Ju and Yuille, Alan and Kortylewski, Adam},
  booktitle={ECCV},
  pages={163--180},
  year={2022}
}

@inproceedings{bento2021timeshap,
  title={Timeshap: Explaining recurrent models through sequence perturbations},
  author={Bento, Jo{\~a}o and Saleiro, Pedro and Cruz, Andr{\'e} F and Figueiredo, M{\'a}rio AT and Bizarro, Pedro},
  booktitle={SIGKDD},
  pages={2565--2573},
  year={2021}
}

@inproceedings{choi2016retain,
  title={Retain: An interpretable predictive model for healthcare using reverse time attention mechanism},
  author={Choi, Edward and Bahadori, Mohammad Taha and Sun, Jimeng and Kulas, Joshua and Schuetz, Andy and Stewart, Walter},
  booktitle={NeurIPS},
pages={3504--3512},
  year={2016}
}

@inproceedings{suresh2017clinical,
  title={Clinical intervention prediction and understanding with deep neural networks},
  author={Suresh, Harini and Hunt, Nathan and Johnson, Alistair and Celi, Leo Anthony and Szolovits, Peter and Ghassemi, Marzyeh},
  booktitle={MLHC},
  pages={322--337},
  year={2017}
}

@inproceedings{tonekaboni2020went, title={What went wrong and when? {I}nstance-wise feature importance for time-series black-box models}, author={Tonekaboni, Sana and Joshi, Shalmali and Campbell, Kieran and Duvenaud, David K and Goldenberg, Anna}, booktitle={NeurIPS}, pages={799--809}, year={2020} }

@inproceedings{liu2024explaining,
      title={Explaining Time Series via Contrastive and Locally Sparse Perturbations}, 
      author={Zichuan Liu and Yingying Zhang and Tianchun Wang and Zefan Wang and Dongsheng Luo and Mengnan Du and Min Wu and Yi Wang and Chunlin Chen and Lunting Fan and Qingsong Wen},
      year={2024},
pages={1--21},
      booktitle={ICLR}
}

@inproceedings{enguehard23a,
  title = {Learning Perturbations to Explain Time Series Predictions},
  author =   {Enguehard, Joseph},
  booktitle = {ICML},
  pages = 	 {9329--9342},
  year = 	 {2023}
}

@inproceedings{jang2016categorical,
  title={Categorical Reparameterization with Gumbel-Softmax},
  author={Jang, Eric and Gu, Shixiang and Poole, Ben},
  booktitle={ICLR},
  pages={1--12},
  year={2017}
}

@article{dau2019ucr,
  title={The UCR time series archive},
  author={Dau, Hoang Anh and Bagnall, Anthony and Kamgar, Kaveh and Yeh, Chin-Chia Michael and Zhu, Yan and Gharghabi, Shaghayegh and Ratanamahatana, Chotirat Ann and Keogh, Eamonn},
  journal={IEEE/CAA Journal of Automatica Sinica},
  volume={6},
  number={6},
  pages={1293--1305},
  year={2019},
  publisher={IEEE}
}

@article{kaushik2020ai,
  title={{AI} in healthcare: time-series forecasting using statistical, neural, and ensemble architectures},
  author={Kaushik, Shruti and Choudhury, Abhinav and Sheron, Pankaj Kumar and Dasgupta, Nataraj and Natarajan, Sayee and Pickett, Larry A and Dutt, Varun},
  journal={Frontiers in Big Data},
volume={3},
  pages={4},
  year={2020},
  publisher={Frontiers Media SA}
}

@INPROCEEDINGS{COMTE,
  author={Ates, Emre and Aksar, Burak and Leung, Vitus J. and Coskun, Ayse K.},
  booktitle={ICAPAI}, 
  title={Counterfactual Explanations for Multivariate Time Series}, 
  year={2021},
  volume={},
  number={},
  pages={1-8},
  doi={10.1109/ICAPAI49758.2021.9462056}}

@INPROCEEDINGS{MCELS,
  author={Li, Peiyu and Bahri, Omar and Boubrahimi, SoukaÏna Filali and Hamdi, Shah Muhammad},
  booktitle={ICMLA}, 
  title={M-CELS: Counterfactual Explanation for Multivariate Time Series Data Guided by Learned Saliency Maps}, 
  year={2024},
  volume={},
  number={},
  pages={713-718},
  doi={10.1109/ICMLA61862.2024.00103}}

@article{CONFETTI, 
title={Counterfactual eXplainable AI (XAI) Method for Deep Learning-Based Multivariate Time Series Classification}, volume={40}, 
url={https://ojs.aaai.org/index.php/AAAI/article/view/38792}, 
DOI={10.1609/aaai.v40i21.38792}, 
abstractNote={Recent advances in deep learning have improved multivariate time series (MTS) classification and regression by capturing complex patterns, but their lack of transparency hinders decision-making. Explainable AI (XAI) methods offer partial insights, yet often fall short of conveying the full decision space. Counterfactual Explanations (CE) provide a promising alternative, but current approaches typically prioritize either accuracy, proximity or sparsity -- rarely all -- limiting their practical value. To address this, we propose CONFETTI, a novel multi-objective CE method for MTS. CONFETTI identifies key MTS subsequences, locates a counterfactual target, and optimally modifies the time series to balance prediction confidence, proximity and sparsity. This method provides actionable insights with minimal changes, improving interpretability, and decision support. CONFETTI is evaluated on seven MTS datasets from the UEA archive, demonstrating its effectiveness in various domains. CONFETTI consistently outperforms state-of-the-art CE methods in its optimization objectives, and in six other metrics from the literature, achieving ≥ 10% higher confidence while improving sparsity in ≥ 40%.}, 
number={21}, 
journal={AAAI}, 
author={Cetina, Alan Gabriel Paredes and Benguessoum, Kaouther and Lourenco, Raoni and Kubler, Sylvain}, 
year={2026}, 
month={Mar.}, 
pages={17393-17400} 
}

@inproceedings{li2023attention, 
title={Attention-Based Counterfactual Explanation for Multivariate Time Series}, 
author={Li, Peiyu and Bahri, Omar and Boubrahimi, Souka{"\i}na Filali and Hamdi, Shah Muhammad}, 
booktitle={International Conference on Big Data Analytics and Knowledge Discovery}, 
pages={287--293}, 
year={2023}, 
organization={Springer} }

@InProceedings{MGCF,
author="Li, Peiyu
and Boubrahimi, Souka{\"i}na Filali
and Hamdi, Shah Muhammad",
editor="Rousseau, Jean-Jacques
and Kapralos, Bill",
title="Motif-Guided Time Series Counterfactual Explanations",
booktitle="ICPR",
year="2023",
publisher="Springer Nature Switzerland",
address="Cham",
pages="203--215",
isbn="978-3-031-37731-0"
}

@inproceedings{wang2021learning,
  title={Learning time series counterfactuals via latent space representations},
  author={Wang, Zhendong and Samsten, Isak and Mochaourab, Rami and Papapetrou, Panagiotis},
  booktitle={International Conference on Discovery Science},
  pages={369--384},
  year={2021},
  organization={Springer}
}

@article{bahri2022shapelet,
  title={Shapelet-based counterfactual explanations for multivariate time series},
  author={Bahri, Omar and Boubrahimi, Soukaina Filali and Hamdi, Shah Muhammad},
  journal={arXiv preprint arXiv:2208.10462},
  year={2022}
}

@article{wang2024glacier,
  title={Glacier: guided locally constrained counterfactual explanations for time series classification},
  author={Wang, Zhendong and Samsten, Isak and Miliou, Ioanna and Mochaourab, Rami and Papapetrou, Panagiotis},
  journal={Machine Learning},
  volume={113},
  number={7},
  pages={4639--4669},
  year={2024},
  publisher={Springer}
}

@inproceedings{hollig2022tsevo,
  title={Tsevo: Evolutionary counterfactual explanations for time series classification},
  author={H{\"o}llig, Jacqueline and Kulbach, Cedric and Thoma, Steffen},
  booktitle={ICMLA},
  pages={29--36},
  year={2022},
  organization={IEEE}
}

@inproceedings{refoyo2024sub,
  title={Sub-space: Subsequence-based sparse counterfactual explanations for time series classification problems},
  author={Refoyo, Mario and Luengo, David},
  booktitle={World Conference on Explainable Artificial Intelligence},
  pages={3--17},
  year={2024},
  organization={Springer}
}

@inproceedings{slack2020fooling,
author = {Slack, Dylan and Hilgard, Sophie and Jia, Emily and Singh, Sameer and Lakkaraju, Himabindu},
title = {Fooling LIME and SHAP: Adversarial Attacks on Post hoc Explanation Methods},
year = {2020},
isbn = {9781450371100},
publisher = {Association for Computing Machinery},
address = {New York, NY, USA},
url = {https://doi.org/10.1145/3375627.3375830},
doi = {10.1145/3375627.3375830},
booktitle = {AIES},
pages = {180–186},
numpages = {7},
location = {New York, NY, USA},
series = {AIES '20}
}

@inproceedings{laugel2019dangers,
author = {Laugel, Thibault and Lesot, Marie-Jeanne and Marsala, Christophe and Renard, Xavier and Detyniecki, Marcin},
title = {The dangers of post-hoc interpretability: unjustified counterfactual explanations},
year = {2019},
isbn = {9780999241141},
publisher = {AAAI Press},
booktitle = {IJCAI},
pages = {2801–2807},
numpages = {7},
location = {Macao, China},
series = {IJCAI'19}
}

@inproceedings{freiesleben2022magic,
author = {Kommiya Mothilal, Ramaravind and Mahajan, Divyat and Tan, Chenhao and Sharma, Amit},
title = {Towards Unifying Feature Attribution and Counterfactual Explanations: Different Means to the Same End},
year = {2021},
isbn = {9781450384735},
publisher = {Association for Computing Machinery},
address = {New York, NY, USA},
url = {https://doi.org/10.1145/3461702.3462597},
doi = {10.1145/3461702.3462597},
booktitle = {AIES},
pages = {652–663},
numpages = {12},
location = {Virtual Event, USA},
series = {AIES '21}
}

@article{chen2026signals,
  title={From Signals to Semantics: A Survey on Time Series Explainability through a Human-Cognitive Lens},
  author={Chen, Zhuomin and Lucchesi, Gabriel and Dong, Qingkai and Zheng, Xu and Song, Dongjin and Wen, Qingsong and Cheng, Wei and Ni, Jingchao and Luo, Dongsheng},
  journal={Authorea Preprints},
  year={2026},
  publisher={Authorea}
}

@inproceedings{mothilal2021towards,
author = {Kommiya Mothilal, Ramaravind and Mahajan, Divyat and Tan, Chenhao and Sharma, Amit},
title = {Towards Unifying Feature Attribution and Counterfactual Explanations: Different Means to the Same End},
year = {2021},
isbn = {9781450384735},
publisher = {Association for Computing Machinery},
address = {New York, NY, USA},
url = {https://doi.org/10.1145/3461702.3462597},
doi = {10.1145/3461702.3462597},
booktitle = {AIES},
pages = {652–663},
numpages = {12},
location = {Virtual Event, USA},
series = {AIES '21}
}

@InProceedings{janzing2020feature,
  title = 	 {Feature relevance quantification in explainable AI: A causal problem},
  author =       {Janzing, Dominik and Minorics, Lenon and Bloebaum, Patrick},
  booktitle = 	 {AISTATS},
  pages = 	 {2907--2916},
  year = 	 {2020},
  editor = 	 {Chiappa, Silvia and Calandra, Roberto},
  volume = 	 {108},
  series = 	 {Proceedings of Machine Learning Research},
  month = 	 {26--28 Aug},
  publisher =    {PMLR},
  url = 	 {https://proceedings.mlr.press/v108/janzing20a.html}
}

@article{miller2019explanation,
title = {Explanation in artificial intelligence: Insights from the social sciences},
journal = {Artificial Intelligence},
volume = {267},
pages = {1-38},
year = {2019},
issn = {0004-3702},
doi = {https://doi.org/10.1016/j.artint.2018.07.007},
url = {https://www.sciencedirect.com/science/article/pii/S0004370218305988},
author = {Tim Miller}
}

@InProceedings{mcallester2020formal,
  title = 	 {Formal Limitations on the Measurement of Mutual Information},
  author =       {McAllester, David and Stratos, Karl},
  booktitle = 	 {AISTATS},
  pages = 	 {875--884},
  year = 	 {2020},
  editor = 	 {Chiappa, Silvia and Calandra, Roberto},
  volume = 	 {108},
  series = 	 {Proceedings of Machine Learning Research},
  month = 	 {26--28 Aug},
  publisher =    {PMLR},
  url = 	 {https://proceedings.mlr.press/v108/mcallester20a.html}
}

@inproceedings{hooker2019benchmark,
 author = {Hooker, Sara and Erhan, Dumitru and Kindermans, Pieter-Jan and Kim, Been},
 booktitle = {NeurIPS},
 editor = {H. Wallach and H. Larochelle and A. Beygelzimer and F. d\textquotesingle Alch\'{e}-Buc and E. Fox and R. Garnett},
 pages = {},
 publisher = {Curran Associates, Inc.},
 title = {A Benchmark for Interpretability Methods in Deep Neural Networks},
 url = {https://proceedings.neurips.cc/paper_files/paper/2019/file/fe4b8556000d0f0cae99daa5c5c5a410-Paper.pdf},
 volume = {32},
 year = {2019}
}

@inproceedings{ICLR2025_2231d1ab,
 author = {Zheng, Xu and Shirani, Farhad and Chen, Zhuomin and Lin, Chaohao and Cheng, Wei and Guo, Wenbo and Luo, Dongsheng},
 booktitle = {ICLR},
 editor = {Y. Yue and A. Garg and N. Peng and F. Sha and R. Yu},
 pages = {12772--12804},
 title = {F-Fidelity: A Robust Framework for Faithfulness Evaluation of Explainable AI},
 url = {https://proceedings.iclr.cc/paper_files/paper/2025/file/2231d1abdcc68d16c9861a58f3c25849-Paper-Conference.pdf},
 volume = {2025},
 year = {2025}
}

@article{parzen1962estimation,
  title={On estimation of a probability density function and mode},
  author={Parzen, Emanuel},
  journal={The Annals of Mathematical Statistics},
  volume={33},
  number={3},
  pages={1065--1076},
  year={1962},
  publisher={JSTOR}
}

@inproceedings{roth2022out,
  title={Out-of-distribution detection using union of 1-dimensional subspaces},
  author={Roth, Karsten and others},
  booktitle={CVPR},
  pages={9430--9440},
  year={2022}
}

@article{gretton2012kernel,
  title={A kernel two-sample test},
  author={Gretton, Arthur and Borgwardt, Karsten M and Rasch, Malte J and Sch{\"o}lkopf, Bernhard and Smola, Alexander},
  journal={The journal of machine learning research},
  volume={13},
  number={1},
  pages={723--773},
  year={2012},
  publisher={JMLR. org}
}

@inproceedings{rabanser2019failing,
 author = {Rabanser, Stephan and G\"{u}nnemann, Stephan and Lipton, Zachary},
 booktitle = {NeurIPS},
 editor = {H. Wallach and H. Larochelle and A. Beygelzimer and F. d\textquotesingle Alch\'{e}-Buc and E. Fox and R. Garnett},
 pages = {},
 publisher = {Curran Associates, Inc.},
 title = {Failing Loudly: An Empirical Study of Methods for Detecting Dataset Shift},
 url = {https://proceedings.neurips.cc/paper_files/paper/2019/file/846c260d715e5b854ffad5f70a516c88-Paper.pdf},
 volume = {32},
 year = {2019}
}

@article{kullback1951information,
  title={On information and sufficiency},
  author={Kullback, Solomon and Leibler, Richard A},
  journal={The Annals of Mathematical Statistics},
  volume={22},
  number={1},
  pages={79--86},
  year={1951},
  publisher={JSTOR}
}

@inproceedings{zheng2026uncovering,
title={Uncovering Insights of Compound Flooding with Data-Driven {AI}},
author={Xu Zheng and Chaohao Lin and Sipeng Chen and Zhuomin Chen and Jimeng Shi and Jayantha Obeysekera and Jingchao Ni and Wei Cheng and Jason Liu and Dongsheng Luo},
booktitle={SIGKDD},
year={2026},
url={https://openreview.net/forum?id=CKSPm2zgjs}
}

@article{rudin2019stop,
  title={Stop explaining black box machine learning models for high stakes decisions and use interpretable models instead},
  author={Rudin, Cynthia},
  journal={Nature machine intelligence},
  volume={1},
  number={5},
  pages={206--215},
  year={2019},
  publisher={Nature Publishing Group UK London}
}

@article{ghassemi2021false,
  title={The false hope of current approaches to explainable artificial intelligence in health care},
  author={Ghassemi, Marzyeh and Oakden-Rayner, Luke and Beam, Andrew L},
  journal={The lancet digital health},
  volume={3},
  number={11},
  pages={e745--e750},
  year={2021},
  publisher={Elsevier}
}

@inproceedings{ribeiro2016should,
author = {Ribeiro, Marco Tulio and Singh, Sameer and Guestrin, Carlos},
title = {"Why Should I Trust You?": Explaining the Predictions of Any Classifier},
year = {2016},
isbn = {9781450342322},
publisher = {Association for Computing Machinery},
address = {New York, NY, USA},
url = {https://doi.org/10.1145/2939672.2939778},
doi = {10.1145/2939672.2939778},
booktitle = {SIGKDD},
pages = {1135–1144},
numpages = {10},
location = {San Francisco, California, USA},
series = {KDD '16}
}

@inproceedings{ismail2020benchmarking,
 author = {Ismail, Aya Abdelsalam and Gunady, Mohamed and Corrada Bravo, Hector and Feizi, Soheil},
 booktitle = {NeurIPS},
 editor = {H. Larochelle and M. Ranzato and R. Hadsell and M.F. Balcan and H. Lin},
 pages = {6441--6452},
 publisher = {Curran Associates, Inc.},
 title = {Benchmarking Deep Learning Interpretability in Time Series Predictions},
 url = {https://proceedings.neurips.cc/paper_files/paper/2020/file/47a3893cc405396a5c30d91320572d6d-Paper.pdf},
 volume = {33},
 year = {2020}
}

@InProceedings{pmlr-v151-pawelczyk22a,
  title = 	 { Exploring Counterfactual Explanations Through the Lens of Adversarial Examples: A Theoretical and Empirical Analysis },
  author =       {Pawelczyk, Martin and Agarwal, Chirag and Joshi, Shalmali and Upadhyay, Sohini and Lakkaraju, Himabindu},
  booktitle = 	 {AISTATS},
  pages = 	 {4574--4594},
  year = 	 {2022},
  editor = 	 {Camps-Valls, Gustau and Ruiz, Francisco J. R. and Valera, Isabel},
  volume = 	 {151},
  series = 	 {Proceedings of Machine Learning Research},
  month = 	 {28--30 Mar},
  publisher =    {PMLR},
  url = 	 {https://proceedings.mlr.press/v151/pawelczyk22a.html}
}

@InProceedings{pmlr-v180-nemirovsky22a,
  title = 	 {CounteRGAN: Generating counterfactuals for real-time recourse and interpretability using residual GANs},
  author =       {Nemirovsky, Daniel and Thiebaut, Nicolas and Xu, Ye and Gupta, Abhishek},
  booktitle = 	 {UAI},
  pages = 	 {1488--1497},
  year = 	 {2022},
  editor = 	 {Cussens, James and Zhang, Kun},
  volume = 	 {180},
  series = 	 {Proceedings of Machine Learning Research},
  month = 	 {01--05 Aug},
  publisher =    {PMLR},
  url = 	 {https://proceedings.mlr.press/v180/nemirovsky22a.html}
}

@inproceedings{dice,
author = {Mothilal, Ramaravind K. and Sharma, Amit and Tan, Chenhao},
title = {Explaining machine learning classifiers through diverse counterfactual explanations},
year = {2020},
isbn = {9781450369367},
publisher = {Association for Computing Machinery},
address = {New York, NY, USA},
url = {https://doi.org/10.1145/3351095.3372850},
doi = {10.1145/3351095.3372850},
booktitle = {Proceedings of the 2020 Conference on Fairness, Accountability, and Transparency},
pages = {607–617},
numpages = {11},
location = {Barcelona, Spain},
series = {FAT* '20}
}

@InProceedings{Jeanneret_2022_ACCV,
    author    = {Jeanneret, Guillaume and Simon, Loic and Jurie, Frederic},
    title     = {Diffusion Models for Counterfactual Explanations},
    booktitle = {ACCV},
    month     = {December},
    year      = {2022},
    pages     = {858-876}
}

\end{document}